\documentclass[11pt]{article}

\usepackage[preprint]{acl}

\usepackage{times}
\usepackage{latexsym}

\usepackage[T1]{fontenc}

\usepackage[utf8]{inputenc}

\usepackage{microtype}

\usepackage{inconsolata}

\usepackage{graphicx}

\usepackage{makecell}  
\usepackage{algorithm}
\usepackage{algpseudocode}
\usepackage{amsmath}
\usepackage{adjustbox}
\usepackage{booktabs}
\usepackage{tabularx}
\usepackage{array}
\usepackage{xcolor}

\usepackage{enumitem}

\newcolumntype{Y}{>{\centering\arraybackslash}X} 
\newcommand{\tallrow}{\rule{0pt}{5.5ex}}    

\newcommand{\dfraccell}[1]{\(\displaystyle #1\)}

\title{SSSD: Simply-Scalable Speculative Decoding}

\author{
 \textbf{Michele Marzollo\textsuperscript{1}},
 \textbf{Jiawei Zhuang\textsuperscript{1}},
 \textbf{Niklas Roemer\textsuperscript{1,2}},
 \textbf{Niklas Zwingenberger\textsuperscript{1}},
\\
 \textbf{Lorenz K. Muller\textsuperscript{1}},
 \textbf{Lukas Cavigelli\textsuperscript{1}}
\\
\\
 \textsuperscript{1}Huawei,
 \textsuperscript{2}ETH Zurich
\\
 \small{
   \textbf{Correspondence:} \href{mailto:michele.marzollo@huawei.com}{michele.marzollo@huawei.com}
 }
}

\begin{document}
\maketitle
\begin{abstract}
Speculative decoding has emerged as a popular technique for accelerating inference in Large Language Models. However, most existing approaches yield only modest improvements in production serving systems. Methods that achieve substantial speedups typically rely on an additional trained draft model or auxiliary model components, increasing deployment and maintenance complexity. This added complexity reduces flexibility, particularly when serving workloads shift to tasks, domains, or languages that are not well represented in the draft model's training data. \\
We introduce Simply-Scalable Speculative Decoding (SSSD), a training-free method that combines lightweight n-gram matching with hardware-aware speculation. Relative to standard autoregressive decoding, SSSD reduces latency by up to 2.9×. It achieves performance on par with leading training-based approaches across a broad range of benchmarks, while requiring substantially lower adoption effort---no data preparation, training or tuning are needed---and exhibiting superior robustness under language and domain shift, as well as in long-context settings.
\end{abstract}

\section{Introduction}

Large Language Models (LLMs) have become ubiquitous across a wide variety of applications, yet their high memory and compute requirements continue to pose substantial challenges for delivering low-latency, cost-efficient user experiences. LLM inference consists of two main phases: the prefill phase, where the input sequence is processed in parallel to compute and store the KV-cache and generate the first token, and the autoregressive decoding phase, where the model generates one token at a time through a full pass over the model weights \citep{yuan2024llminferenceunveiledsurvey}.

Speculative decoding (SD) \citep{xia-etal-2023-speculative,leviathan2023,chen2023acceleratinglargelanguagemodel} breaks the autoregressive loop by using a smaller model to ``guess'' future tokens (the \textit{drafting} phase), which are then verified in a single forward pass of the target model. If some guesses are correct, a single iteration produces multiple tokens and the overall latency is reduced while maintaining the exact output distribution of the original model. In practice, real-world adoption has remained limited: suitable drafting models are often unavailable and the cost of training a well-aligned draft model can be prohibitive \cite{li2024eagle}. Despite the emergence of numerous SD variants, many approaches fail to deliver speedups at practical batch sizes, while introducing substantial implementation and deployment complexity. In production systems, where ease of integration and cost efficiency matter as much as raw latency improvements, these limitations often outweigh the gains.

In this paper, we present a new approach explicitly designed for practical deployment. Our method uses a lightweight, custom-built n-gram model that runs entirely on CPU. It keeps adoption simple by avoiding the training or deployment of any additional models, enabling simpler integration into existing systems and better generalization to new domains, tasks, or languages without manual intervention or additional engineering effort. The source code is available at \url{https://github.com/huawei-csl/sssd_speculator}.

\section{Background and Related Work} \label{sec:related-work}

Since the introduction of SD, numerous variants have been proposed. While a detailed overview of all existing methods is beyond the scope of this work, \citet{xia-etal-2024-unlocking} provide an extensive survey of the field.

\subsection{Main speculative decoding methods}

The standard approach to SD uses a smaller, faster draft model to autoregressively generate candidate tokens for verification. For this approach to be effective, the draft model must be closely aligned with the target model, as generating candidates incurs nontrivial computational cost.
Typically, effective draft models are drawn from the same model family as the target model, sharing training data and recipes. However, such models are often unavailable, and training new ones can be prohibitively expensive. To address this limitation, \citet{spector2023acceleratingllminferencestaged} propose tree-based speculation, which generates multiple possible continuations rather than a single sequence, thereby increasing the number of accepted tokens, as well as staged speculation, which applies speculative decoding recursively to the draft model itself. Medusa \citep{cai2024medusa} replaces the draft model with multiple decoding heads attached to the target model’s final layer. These heads can be trained through self-distillation using substantially less data, preserving alignment with the target distribution while reducing training costs. Similarly, EAGLE \citep{li2024eagle,li2024eagle2} trains a single autoregressive head to generate candidate trees. EAGLE-3 \citep{li2025eagle3} further advances this method by leveraging hidden states from multiple layers of the target model as input for the speculative head, achieving state-of-the-art inference speedups.

All these methods require training draft models or heads, deploying them on accelerators---which can be challenging in distributed settings \citep{chen2023acceleratinglargelanguagemodel}---and retraining whenever the target model or application domain changes \citep{hong2025training}.
Multi-token prediction (MTP) \citep{gloeckle2024mtp} introduces additional heads into the target model at training time, modifying the classical next-token prediction (NTP) training objective. Although MTP is primarily designed to improve model quality, these auxiliary heads can also be leveraged for SD. Nevertheless, aside from a few notable examples \citep{deepseekai2025deepseekv3technicalreport}, this paradigm has not yet seen widespread adoption.

\subsection{Multilinguality}

Prior work has shown that the speculation accuracy of draft models varies substantially across languages. \citet{sandler2025disparateimpactsspeculativedecoding} propose a fine-tuning strategy to mitigate speedup disparities across underrepresented tasks and languages. \citet{yi-etal-2024-towards} observe that small draft models, such as Medusa and EAGLE, perform well in English but generalize poorly to other languages due to their limited capacity, and propose training language-specific drafters. While these approaches effectively address important limitations of speculative decoding, they do so at the cost of increased deployment and maintenance complexity, which further motivates a training-free, self-adaptable alternative.

\subsection{Model-free methods} \label{subsec:param-free}

Model-free approaches eliminate the need for additional trainable parameters for drafting and focus instead on lightweight, easily deployable alternatives. Some approaches leverage the target model itself to generate candidate tokens, such as by skipping intermediate layers during the drafting phase \citep{zhang-etal-2024-draft} or by generating a pool of n-grams for use in subsequent speculative iterations \citep{fu2024break}. While simplifying deployment, these approaches come with major computational costs for the drafting phase, which overshadow the speedups given by accepting more tokens. Other model-free methods generate candidate tokens through token-level pattern matching or external retrieval, avoiding reliance on the model itself. Prompt Lookup Decoding (PLD)~\citep{saxena2023prompt}, LLMA \citep{yang2023inferencereferencelosslessacceleration}, and Lookahead~\citep{zhao2023lookahead} look for recent matches in the prompt or across previous generations to propose likely continuations. REST~\citep{he2023rest} and ANPD~\citep{ou2024lossless} use large external corpora to statistically estimate probable next tokens.

These retrieval-based methods are attractive due to their ease of deployment; however, this simplicity comes at a cost: they produce candidate tokens with lower acceptance rates and often achieve good speedups only on specific tasks.

\begin{figure*}[ht]
    \centering
    \includegraphics[width=0.95\textwidth]{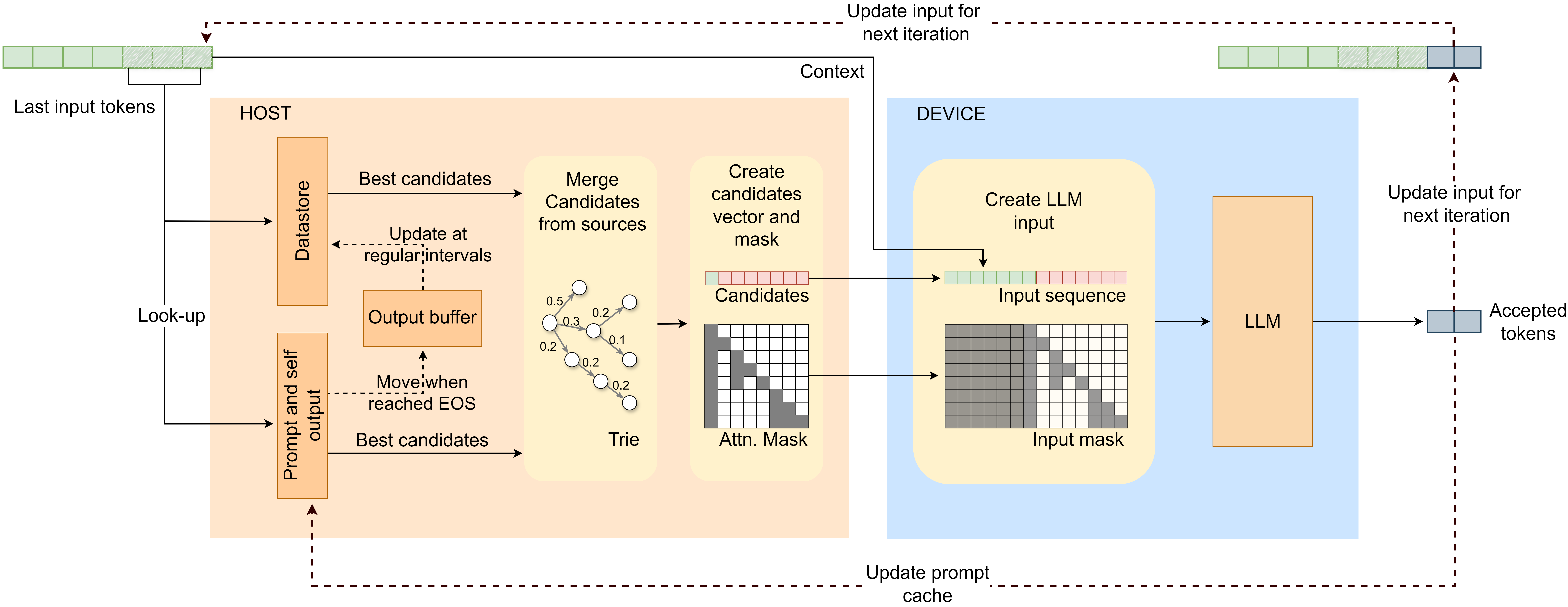}
    \caption{A representation of the system with the main steps of the SSSD method.}
    \label{fig:system_diagram}
\end{figure*}

\subsection{Large-scale speculative decoding}

Most work on SD focuses on accelerating inference for single-prompt scenarios. However, in practical deployments, batching is the primary strategy for reducing cost, which often outweighs latency concerns. While batching limits the effectiveness of speculative techniques due to shared resource constraints \citep{su2023synergyspeculativedecodingbatching}, it remains central to LLM inference systems, which typically employ continuous batching \citep{orca2022} to optimize resource utilization and user experience. \citet{liu2025turbospec} show that, in this context, standard autoregressive decoding can outperform SD at moderate request rates due to the overhead of running the draft model. In contrast, lightweight methods like Prompt Lookup Decoding perform well at high request rates in tasks where simple retrieval is effective.
Nevertheless, a number of model-based approaches report consistent speedups even in batched settings \citep{specinfer2024,zhong2024propd,li2025eagle3,zhang2024recurrent}, in some cases up to batch size 128 in long-context regimes \citep{sadhukhan2025magicdec}.

\paragraph{Contributions}

\begin{enumerate}[leftmargin=*, itemsep=0pt, parsep=0pt]
    \item We propose SSSD, an SD method that requires no data preparation, training, or fine-tuning;
    \item We evaluate and compare the end-to-end latency–throughput trade-off in SGLang;
    \item We demonstrate SSSD’s suitability for task, domain, and language adaptation, including in cold-start settings;
    \item We provide insights into parameter selection, including datastore size, data sources, and hardware-aware speculation length adjustments.
\end{enumerate}

\section{Method}

Designing an efficient and scalable speculative decoding method involves several critical challenges. Most existing work focuses on prediction quality, but two additional aspects are crucial: standard approaches introduce substantial overhead, as the cost of drafting can offset the benefits of correct predictions \citep{liu2025turbospec}, and the verification phase must be optimized to make effective use of hardware resources.

\subsection{Algorithm design} \label{main:algorithm_design}

The first objective of SSSD is to remove the complexity and cost of running the drafting phase on-device (where \textit{device} refers to the GPU or any alternative accelerator). Although skipping model identification, modification, training, and deployment is clearly attractive, existing parameter-free approaches suffer from limited speculation quality.

The core idea of our method is to treat the prompt and self-output as a unified n-gram source and to integrate it with a large text datastore. This datastore can be built from any available corpus or populated automatically during serving by storing outputs from past conversations. Our experiments show that the two sources provide complementary candidates, and that combining them substantially improves draft-token quality.
Figure~\ref{fig:system_diagram} illustrates the process: the final tokens of the input sequence (the draft \textit{prefix}) are matched against both sources, and the continuations of the matches are used to choose the candidates for tree-based verification, as described in the following sections.

\subsubsection{Prompt and self-output} \label{main:input}

We treat the prompt and the model’s previously generated tokens (\textit{self-output}) as a single sequence, which we refer to as the \emph{input}. This input is stored in a trie-like structure that is continuously updated as new tokens are generated, allowing efficient n-gram retrieval.

To select candidate continuations for a prefix of length $P$ (we use $P = 4$), we examine the continuations of all prefix lengths from $1$ to $P$ (see Algorithm~\ref{alg:input_retrieval} in the Appendix). Each of these trees of continuations will be considered when choosing the best tokens to verify.

\subsubsection{Datastore} \label{main:datastore}

The large datastore is constructed as in REST~\citep{he2023rest}, using a set of suffix arrays over tokenized text to efficiently retrieve continuations of a given prefix. Unlike REST, our datastore can be continuously updated with the model’s own outputs. This provides two advantages: (1) the data is aligned with the model’s output distribution, improving speculation quality, and (2) the system can start from an empty datastore, eliminating any need for precomputation. We also modify the datastore query algorithm to improve speculation quality and reduce retrieval time.

For each prefix, we locate its continuations using binary search on each suffix array and then sample a small, fixed number of them at regular intervals. Because the suffix arrays are sorted, interval sampling approximates the underlying distribution. We then build a weighted tree of these continuations, where edge weights represent occurrence counts. If too few continuations are found, we progressively shorten the prefix and repeat the search until reaching a minimum threshold. Continuations are merged at each iteration by summing edge weights (see Algorithm \ref{alg:get_cands} for details). 

Lookahead~\citep{zhao2023lookahead} also combines prompt information with continually updated statistics of the model's output, stored in a tree with aggressive pruning to control memory and retrieval latency. However, pruning degrades candidate quality: rare sequences, though infrequent, collectively account for many useful n-grams. In our experiments, all pruning strategies harmed acceptance rates. Accordingly, for the live datastore we limit size by discarding the oldest entries, avoiding internal pruning. By combining a suffix-array design with an efficient sampling strategy, our approach accommodates significantly more data than previous speculative methods, resulting in higher speculation accuracy while maintaining predictable retrieval cost.

\subsubsection{Datastore management} \label{main:datastore_management}

\begin{figure}[ht]
  \centering
  \includegraphics[width=0.9\columnwidth]{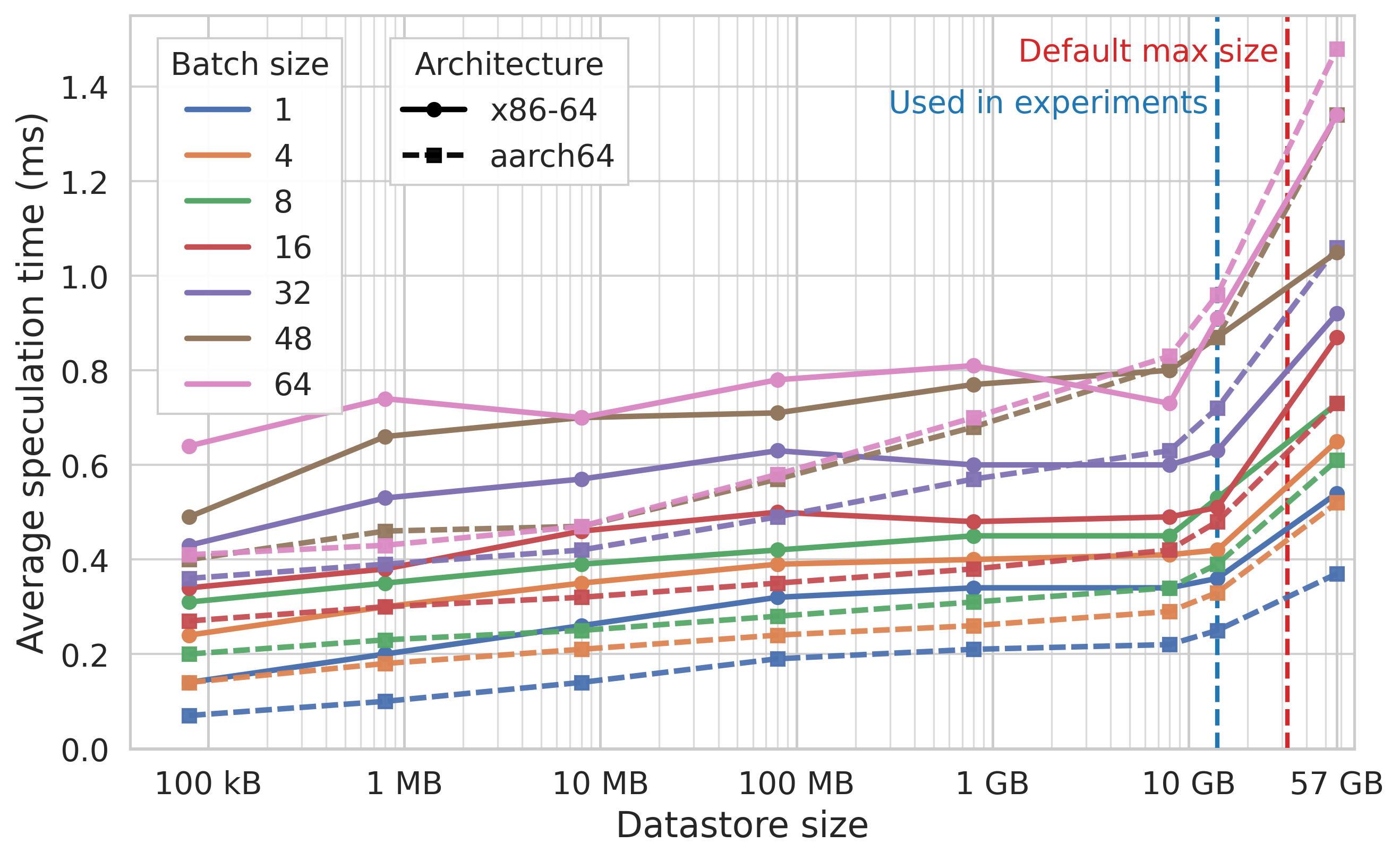}
  \caption{Retrieval cost at increasing datastore sizes and varying batch sizes on x86-64 and aarch64 hosts. The speculation length is fixed at 16 candidates, which is higher than what is typically used at large batch sizes.}
  \label{fig:retrieval_arch}
\end{figure}

As discussed above, the datastore update policy operates over multiple suffix arrays (\textit{sub-indices}) rather than a single index. In the default configuration, each sub-index stores 512M tokens, with 4 bytes for both the value and the index per token ($\approx$4 GB per sub-index). Retrieval samples a fixed number of continuations (100 by default), independent of datastore size, such that adding more sub-indices reduces the number of samples drawn from each.
The number of sub-indices is capped at 8 (typically 4 in our experiments). In the live-datastore setting, when incoming data exceeds this capacity, the oldest sub-index is discarded and replaced with one built from recent data. This design keeps speculation cost bounded and preserves prediction quality by adapting to distribution shifts.

Rebuilding proceeds at regular intervals. If the accumulated data does not fill a sub-index, the last sub-index is rebuilt using its previous content combined with the new data. Otherwise, the last sub-index is filled, a new one is created, and, if the total capacity is exceeded, the oldest is removed once the new one is ready. Rebuilding takes several minutes but runs asynchronously and does not affect retrieval latency; memory pooling avoids allocation and deallocation overhead.

Figure~\ref{fig:retrieval_arch} shows that logarithmic search over the suffix arrays keeps retrieval latency low even for datastores larger than those typically deployed in high-throughput settings. Despite substantial differences in raw CPU performance  (x86-64: 96 cores / 192 threads @ 2.4 GHz; aarch64: 192 cores / 192 threads @ 2.6 GHz), retrieval cost remains low and comparable across both architectures.

\subsubsection{Fusing multiple data sources} \label{main:fusion}

\begin{figure*}[!ht]
    \begin{minipage}{0.68\columnwidth}  
        \centering
        \includegraphics[width=\textwidth]{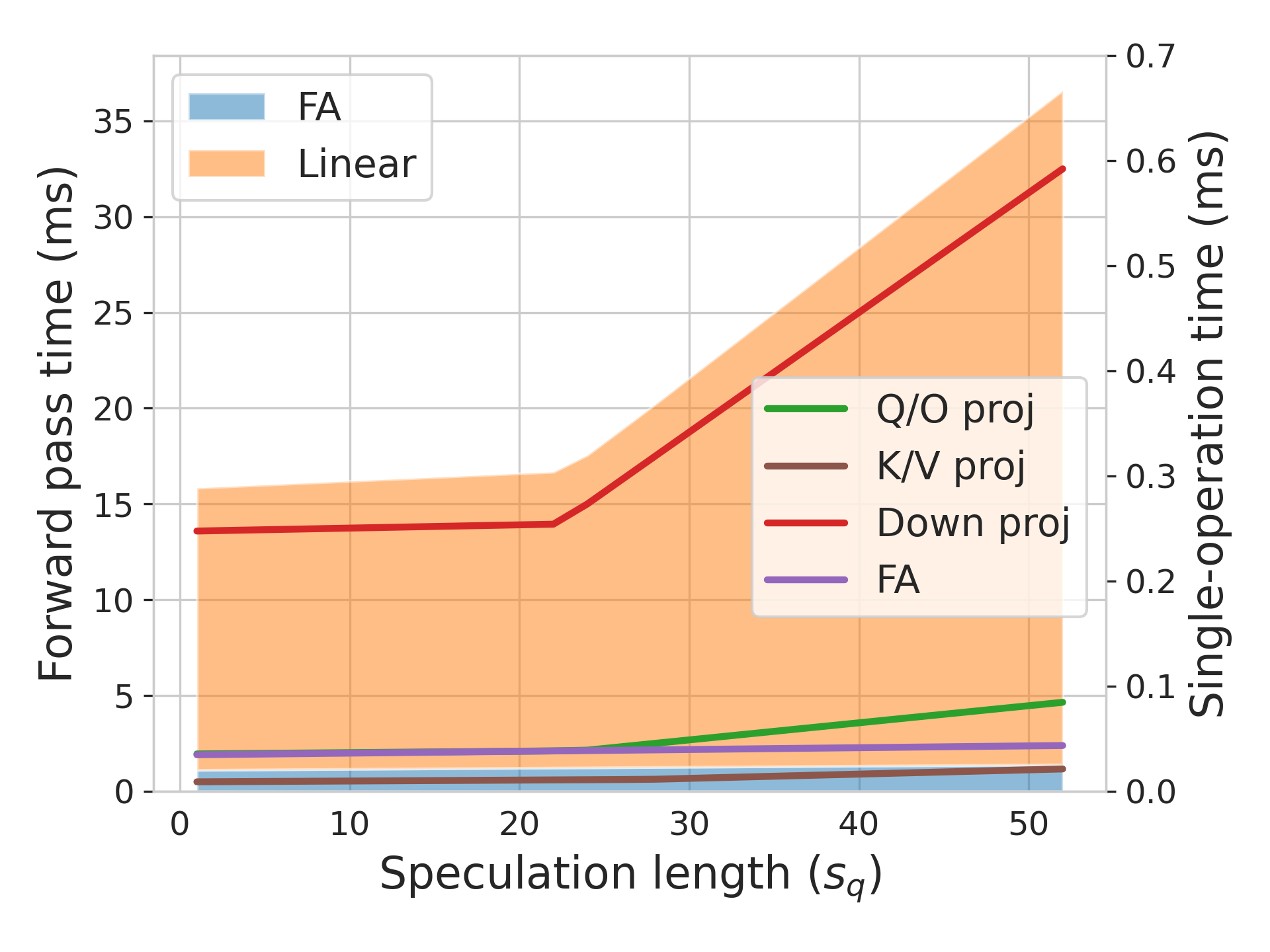}
        {\footnotesize (a) Context length: 1000}
    \end{minipage}
    \begin{minipage}{0.68\columnwidth}  
        \centering
        \includegraphics[width=\textwidth]{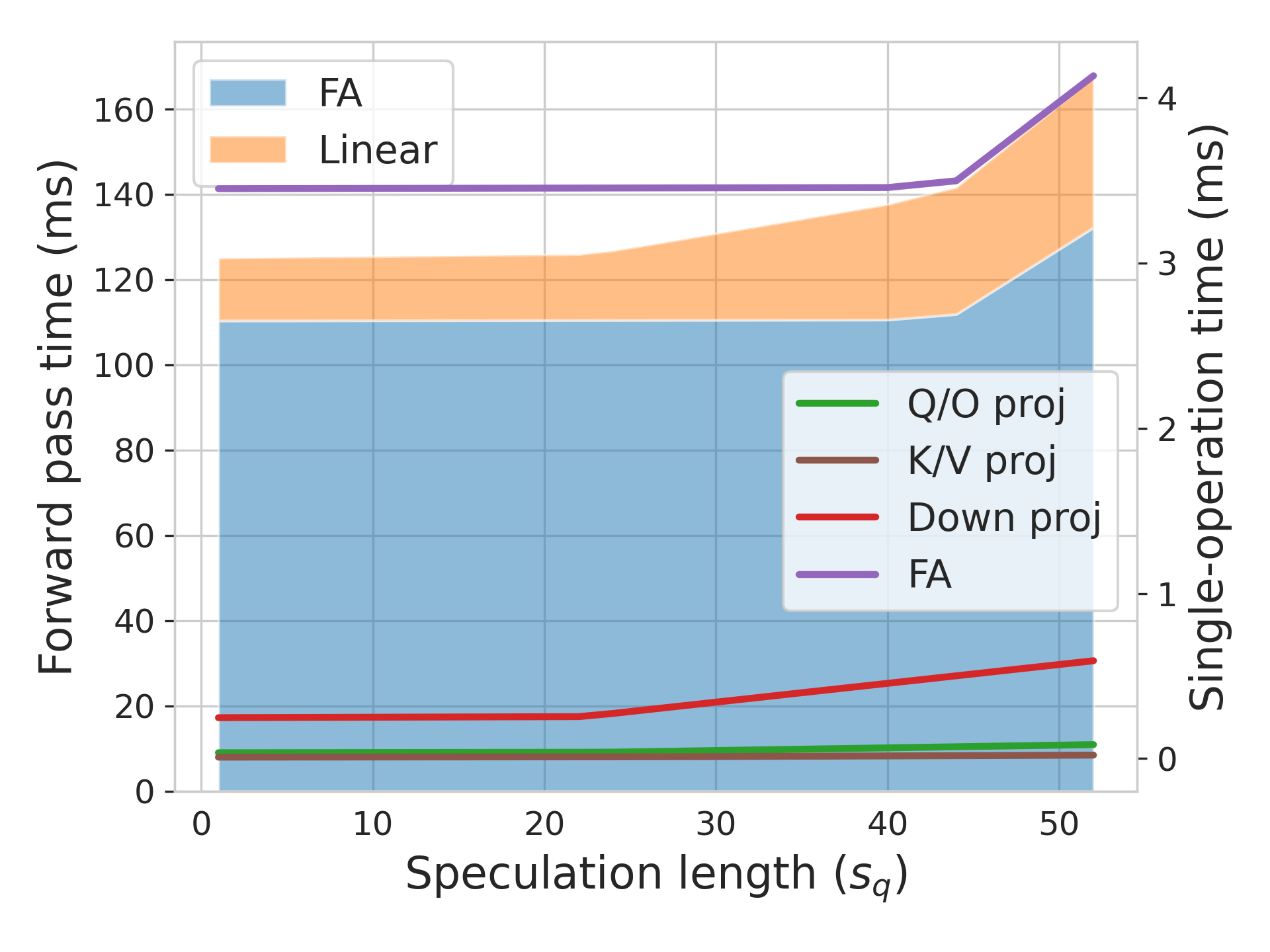}
        {\footnotesize (b) Context length: 100,000}
    \end{minipage}
    \begin{minipage}{0.68\columnwidth}  
        \centering
        \includegraphics[width=\textwidth]{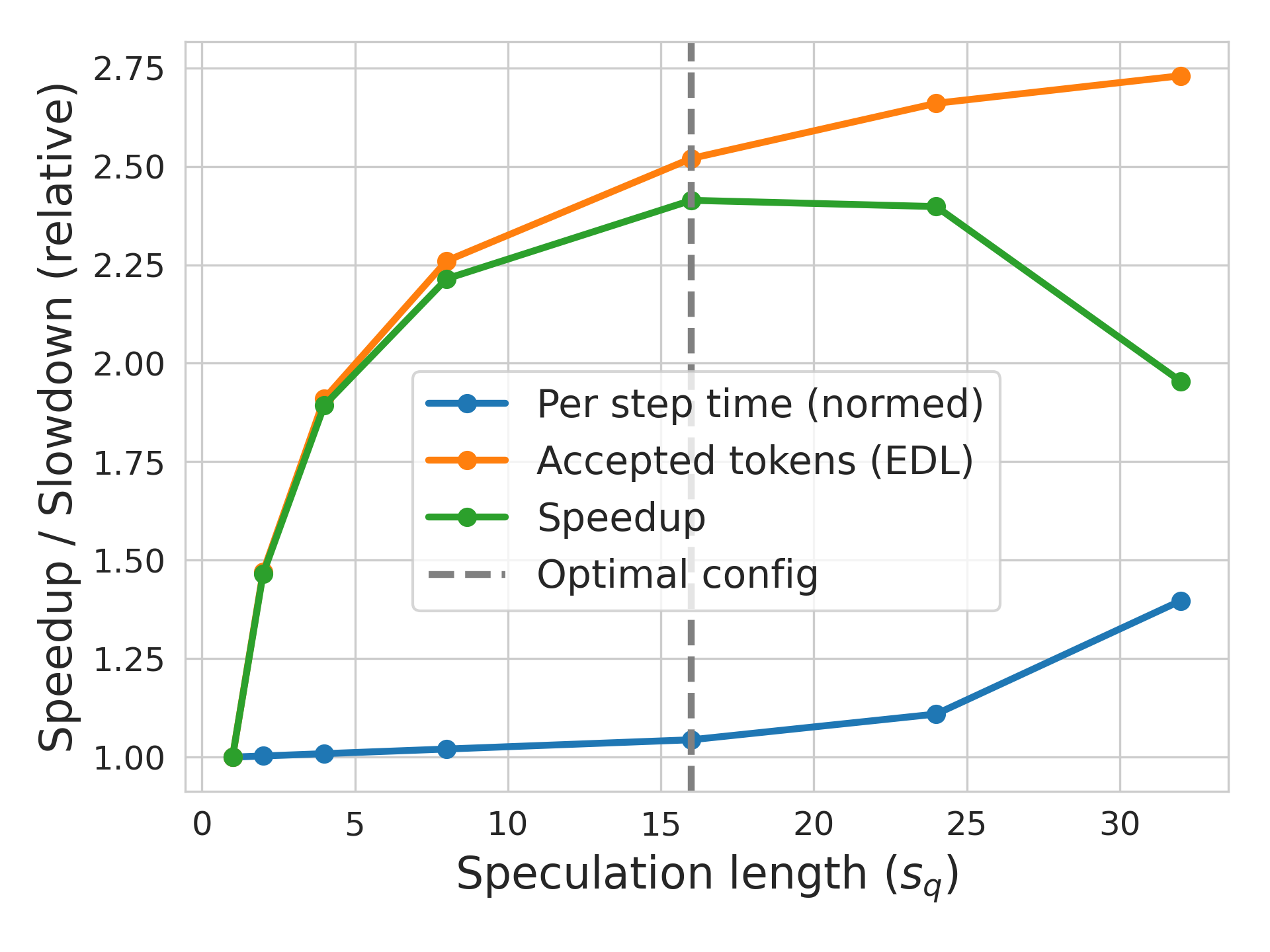}
        {\footnotesize (c) Context length: 1000}
    \end{minipage}
    \caption{(a,b) Roofline-based forward-pass time vs. speculation length, with contributions from linear operations and FlashAttention for Llama-3.1-8B at batch size 8 (same hardware as in Figure~\ref{fig:results_8b}a). Curves correspond to single operations from Table~\ref{tab:formulas}. (c) Accepted tokens, normalized cost, and theoretical speedup vs. speculation length.}
    \label{fig:dissection}
\end{figure*}

After retrieving candidates from the two sources, we must combine them to produce high-quality next-token proposals. Each source provides an estimate of the likelihood that a candidate will be accepted during verification, which we refer to as its \textit{probability}. This pseudo-probability is computed as the number of occurrences of the full sequence (prefix plus candidate tokens) divided by the number of occurrences of the prefix alone. However, these estimates differ in reliability: probabilities derived from the input are typically overconfident compared to those from the datastore. To correct this, we apply a scaling factor to input-based probabilities. Within the input itself, we further dampen probabilities based on the length of the prefix match and the candidate’s depth in the tree. We select these weights once via a simple grid search, and they transfer well across models and tasks.

Appendix~\ref{sec:appendix_algo} provides pseudocode for building the candidate trees and for merging data from multiple sources. From the merged tree, we then construct the candidate vector for verification and the corresponding attention mask.

\subsection{Hardware-aware speculation}
\label{main:roofline_modeling}

Having designed an algorithm that reduces drafting costs and complexity while preserving high speculation accuracy, the final optimization focus is on minimizing the latency overhead of the verification forward pass of the last generated token and the speculated tokens. While many speculative decoding studies highlight the underuse of accelerator resources as an important factor enabling SD, few explore how memory bandwidth and compute resources interact during SD, especially under batching. We propose a simple rule to dynamically choose a near-optimal speculation length based on the current batch size, significantly simplifying serving with SD.

We define the speculation length $s_q$ as the total number of nodes in the speculation tree, including the root (the latest accepted token) and the candidate tokens to be verified. We denote batch size as $b$ and group size for GQA \cite{ainslie2023gqa} as $g$. Figure~\ref{fig:dissection}c illustrates the interaction between speculation length and forward pass time in determining end-to-end acceleration. The orange line represents the average number of tokens accepted by the LLM as a function of $s_q$. The blue line shows the corresponding relative increase in forward step time compared to standard autoregressive decoding ($s_q=1$). Neglecting the drafting time overhead, which remains low and near-constant in our lookup-based approach, the optimal speculation length is the one maximizing the ratio between the algorithmic acceleration (average accepted tokens) and the relative increase in per-step latency.

Appendix~\ref{app:roofline} presents a detailed analysis of the primary device-side operations during a decoding step; here we provide only a simplified overview. Figure~\ref{fig:dissection}a,b illustrates the cost of a forward pass, where the orange region corresponds to the linear layers and the blue region to the attention layers.
In plot (a), where the context is short, the forward-pass cost is dominated by the linear layers. Their cost remains nearly constant as $s_q$ increases, until the computation reaches the compute-bound region of the roofline model at approximately $s_q \approx 22$. Up to this point, speculating $22$ tokens per batch element is almost free.
In Figure~\ref{fig:dissection}b, where the context is long, the dominant cost becomes loading the KV-cache in the attention layers. Although the linear layers follow the same trend as in the short-context case, their relative contribution is smaller, allowing even larger ``free'' speculation budgets.

\begin{figure*}[!ht]
    \centering
    \begin{minipage}{0.68\columnwidth}
        \centering
        \includegraphics[width=\textwidth]{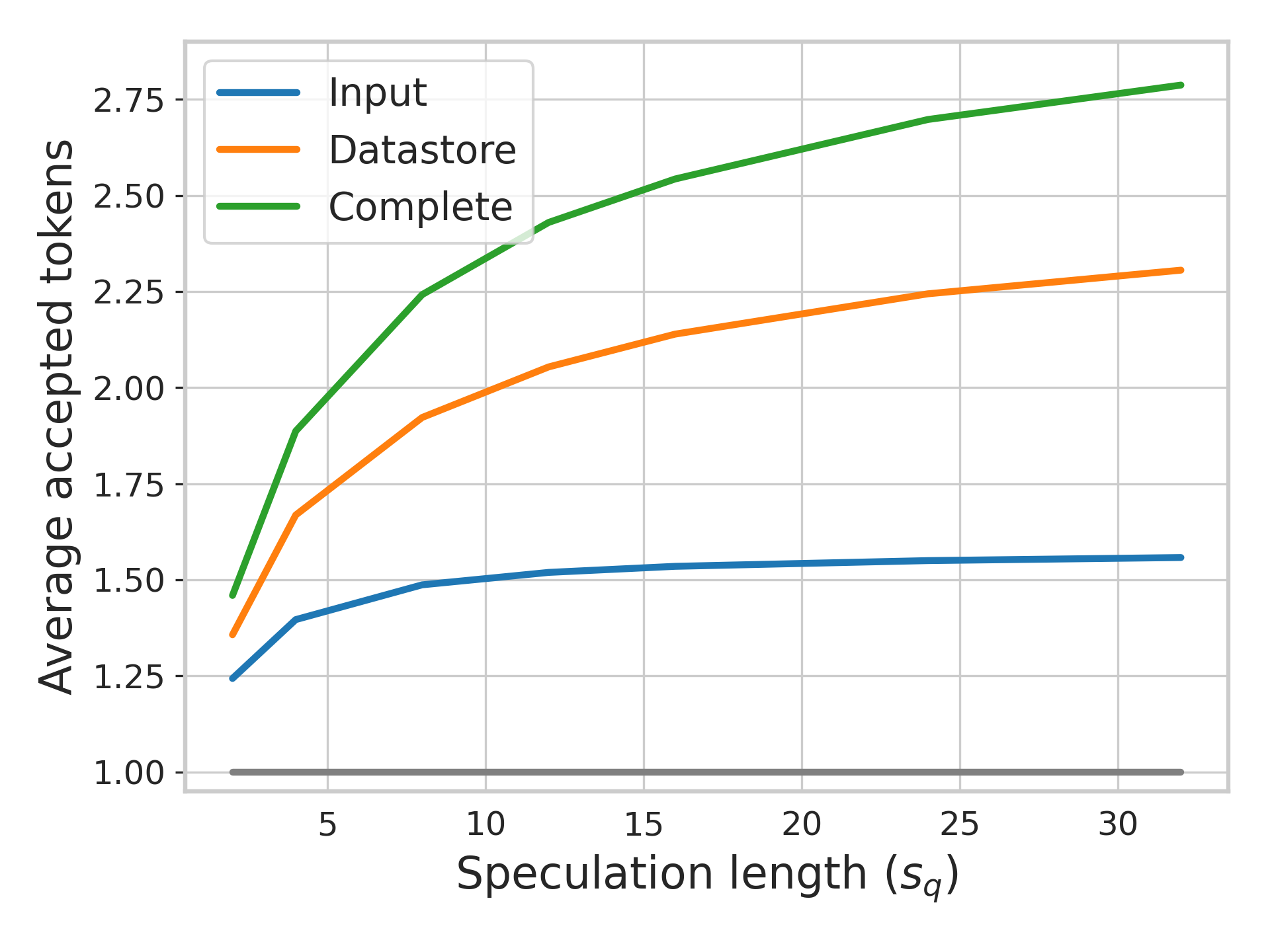}
        {\footnotesize (a) Decomposed components}
    \end{minipage}
    \hfill
    \begin{minipage}{0.68\columnwidth}
        \centering
        \includegraphics[width=\textwidth]{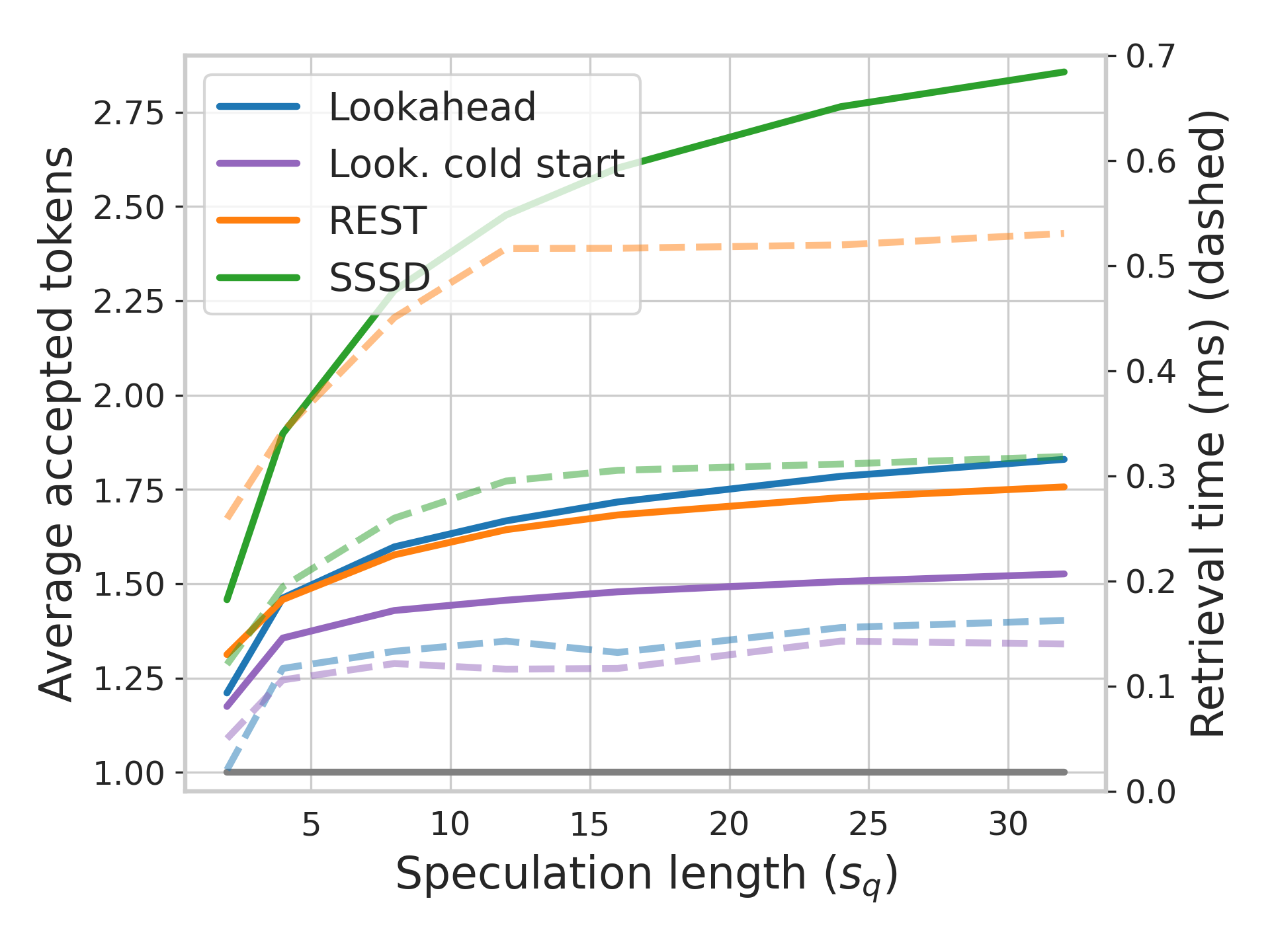}
        {\footnotesize (b) MT-Bench}
    \end{minipage}
    \hfill
    \begin{minipage}{0.68\columnwidth}
        \centering
        \includegraphics[width=\textwidth]{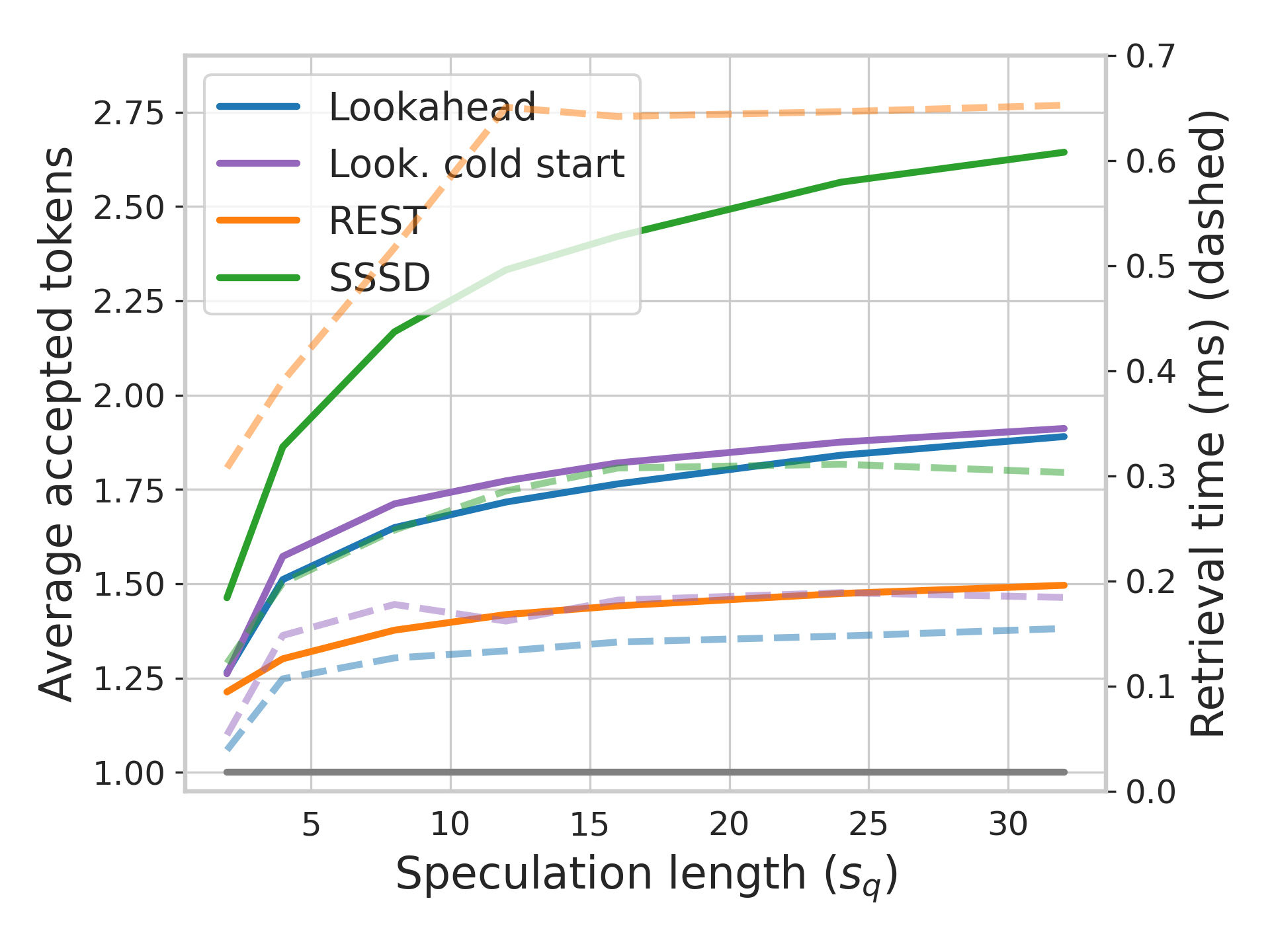}
        {\footnotesize (c) GSM8k}
    \end{minipage}
    
    \caption{(a) Speculation quality of SSSD data sources on 160 MT-Bench and GSM8K prompts. (b,c) Comparison with parameter-free baselines. REST and SSSD use the same datastore; the Lookahead cache is evaluated both warm (identical data as SSSD) and cold. Solid lines show accepted tokens, dashed lines show candidate retrieval and mask construction time. All experiments use Llama-3.1-8B.}
    \label{fig:method_comparisons}
\end{figure*}

As described in the appendix, all operations except attention exhibit a FLOPs-to-IO ratio of approximately $b \times s_q$. 
Following the roofline performance model~\citep{roofline2009}, to operate near the roofline ridge point (i.e., the transition from memory- to compute-bound execution), we set
\[
s_q = \frac{I_\mathrm{knee}}{b},
\]
where $I_\mathrm{knee}$ denotes the arithmetic intensity at the ridge point, defined as the ratio of the accelerator's peak FLOPS to its peak memory bandwidth.

For small batch sizes, the speculation budget implied by this rule can become impractically large for efficient kernel execution, while longer speculation lengths yield diminishing returns. We therefore cap the speculation length to fixed values (e.g., $s_q = 32$ for $b = 1$). This also handles the case where, for $g > b$, the attention operator becomes compute-bound at smaller $s_q$ than the linear layers.

\section{Evaluation}

To evaluate SSSD in a realistic inference setting, we integrate it into SGLang \citep{sglang}, a high-performance LLM inference system. SGLang provides an efficient implementation of EAGLE-2, EAGLE-3 and Lookahead. We further integrate PLD and REST to enable comparisons with leading model-free methods.

Our main experiments focus on models from the Llama-3 family \citep{grattafiori2024llama3herdmodels}, which are widely used in speculative decoding research. For other model families, our choice fell on those with publicly available EAGLE-3 heads.

For EAGLE and Lookahead, we use SGLang’s default parameters for speculation length and tree structure. For other methods, we follow Section~\ref{main:roofline_modeling}. Using SSSD-tuned parameters for EAGLE methods often degrades performance, and exhaustive tuning is computationally challenging and provided limited benefit in preliminary experiments.

\subsection{Algorithm effectiveness}
\label{sec:acceptance_length}

First, we show why our method is particularly effective by breaking down the different components that contribute to selecting candidates for speculation. The method's strength lies in the complementarity of candidates coming from the input and the datastore. This can be seen in Figure \ref{fig:method_comparisons}a, where the contribution of the two main components (the input, which includes prompt and self-output, and the datastore) almost adds up perfectly for large enough $s_q$, meaning that the algorithm balances candidates from the two sources effectively.

We also compare our implementation with the main alternative parameter-free methods, both in terms of speculation quality and retrieval time (see Figure~\ref{fig:method_comparisons}b,c). We use the MT-Bench dataset used in \citep{he2023rest} (the first prompt of each conversation), and the first 80 prompts from the GSM8k dataset \citep{gsm8k}, used in \citep{zhao2023lookahead}, showing that our method consistently outperforms both methods on both datasets. For SSSD and REST, we construct datastores using publicly available data generated either by the same model or models of the same family following the Magpie procedure \citep{xu2025magpie}, and we use the same data to simulate the Lookahead warm-start scenario.

\subsection{Datastore cold-start and multilinguality} \label{sec:multilingual}

\begin{figure*}[!ht]
    \centering
    \begin{minipage}{0.32\textwidth}
        \centering
        \includegraphics[width=\textwidth]{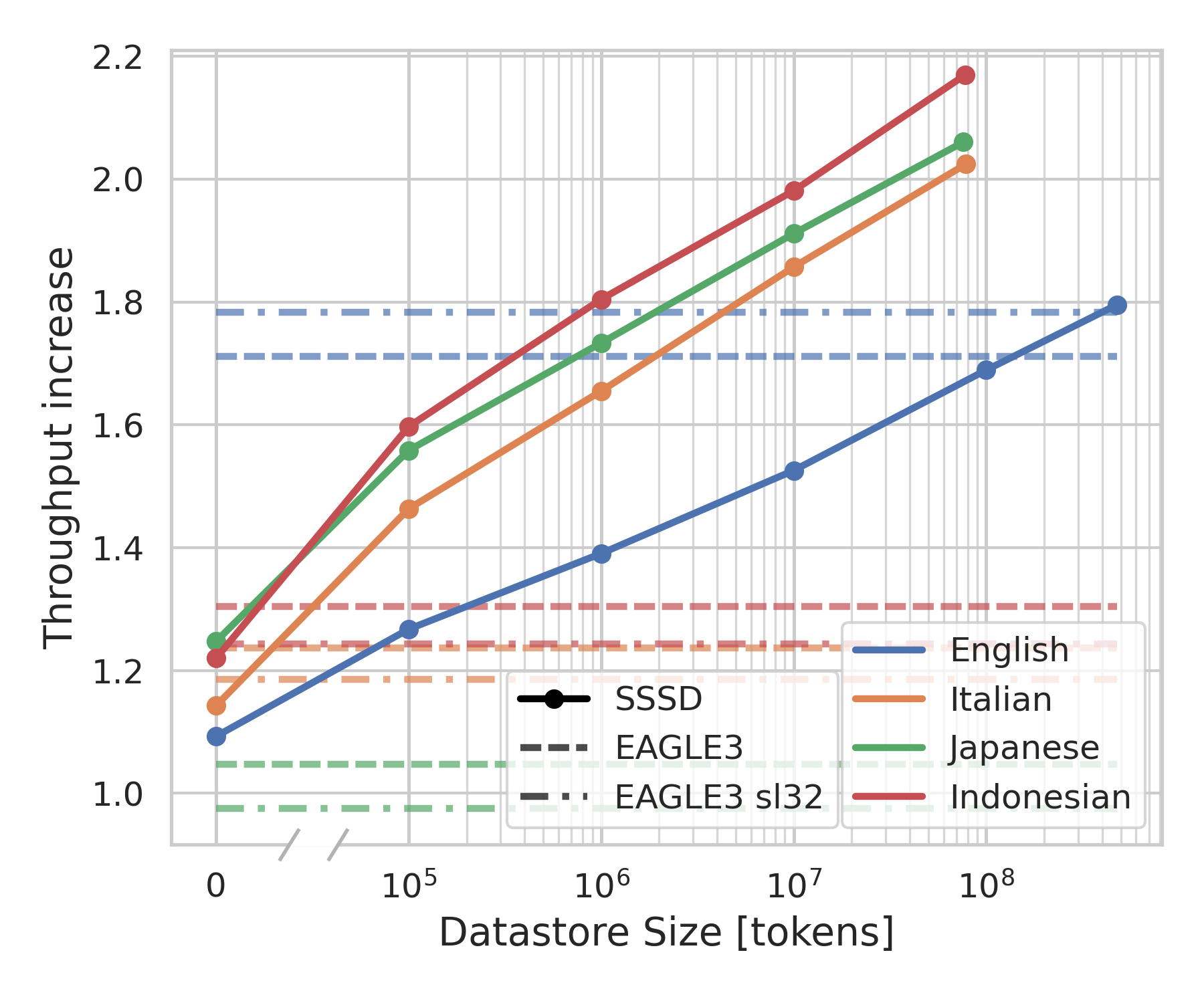}
        {\footnotesize (a) Cold start}
    \end{minipage}
    \hfill
    \begin{minipage}{0.32\textwidth}
        \centering
        \includegraphics[width=\textwidth]{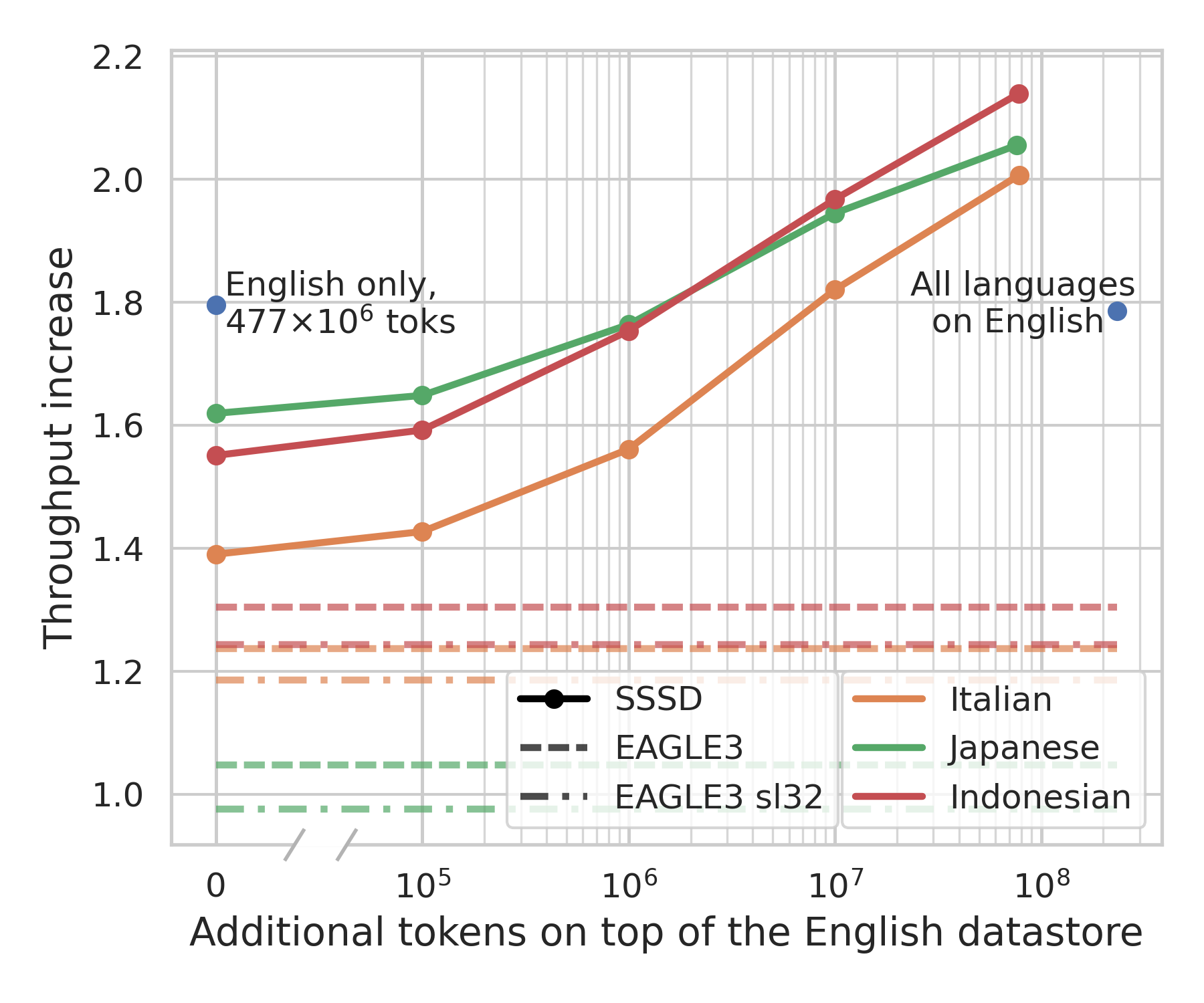}
        {\footnotesize (b) Starting from English datastore}
    \end{minipage}
    \hfill
    \begin{minipage}{0.32\textwidth}
        \centering
        \includegraphics[width=\textwidth]{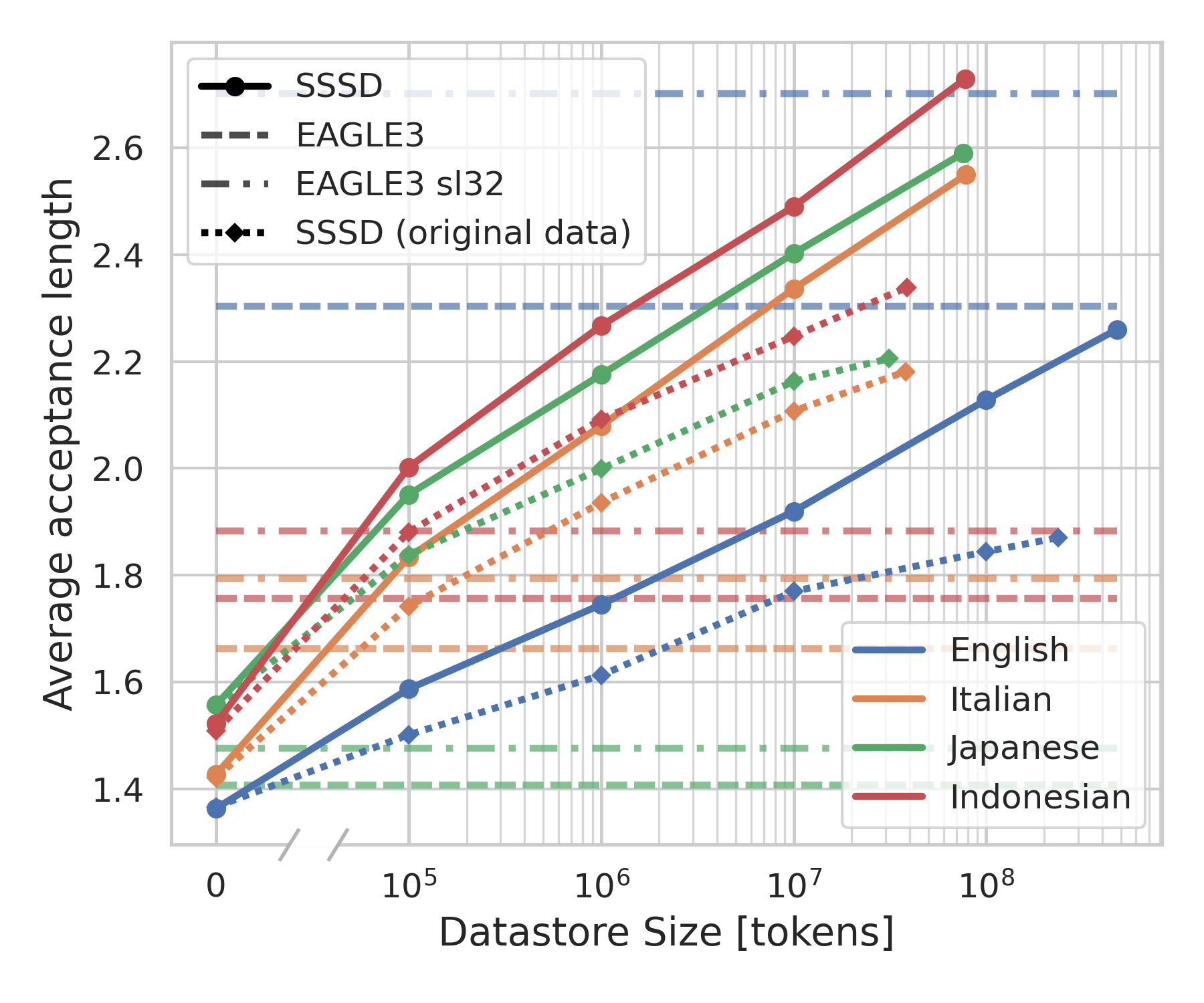}
        {\footnotesize (c) Acceptance lengths (cold start)}
    \end{minipage}

    \caption{Experiments with Qwen3-14B on a 48 GB GPU (165.2 TFLOPS in bfloat16; 1008 GB/s memory bandwidth) on MT-Bench (and translations). Temperature = 0.7, top-p = 0.8, top-k = 20; averages over five runs. (a, b) Speedup over autoregressive decoding at batch size 1 versus datastore size ($\approx$1,000 tokens per user conversation on average). (c) Corresponding acceptance length, comparing model-generated and dataset-derived entries.}
    \label{fig:qwen_speedup}
\end{figure*}

As emphasized earlier, the key strengths of our method lie in its ease of adoption and flexibility across new domains and languages. Accordingly, we investigate its behavior when applied directly to an inference system in a cold-start setting (without a pre-built datastore), examine the differences between using model-generated data and general-domain language data, and evaluate how rapidly SSSD adapts to new languages.

To answer these questions, we use Qwen3-14B \citep{yang2025qwen3technicalreport}, as it offers strong native multilingual support. We evaluate end-to-end speedups as a function of the total number of accumulated tokens in the datastore. Data are generated by the model using prompts from multilingual sources \citep{evol-instruct,sharegpt,freedomintelligence_multilingual,wildchat}. This setup simulates a deployment scenario in which user queries arrive continuously and model outputs are incrementally stored in the datastore.
We measure speedups across a typologically diverse set of languages, chosen to span differences in script, morphological typology, and syntactic structure, including English and three lower-resource languages. We consider two initialization settings: an empty datastore and a large English-only datastore, allowing us to assess both within-language scaling and cross-lingual adaptation.

Our results show that incorporating multiple languages has a negligible negative impact: English performance remains essentially unchanged even when the datastore contains large amounts of non-English data. The method adapts rapidly: after only 1,000 conversations, speedups already reach 1.6×–1.8× for lower-resource languages and approximately 1.4× for English. Importantly, even in a cold-start setting, the model achieves a 1.1×–1.23× speedup, enabling immediate deployment without initial performance degradation. Starting from an English datastore when speculating new languages gives an initial benefit, and tends to converge to the behavior of the monolingual datastore when enough data is generated. Increasing the amount of data is consistently beneficial, yielding approximately logarithmic gains. With sufficient data, SSSD outperforms EAGLE-3 across all evaluated languages, with especially pronounced improvements for lower-resource languages.

In Figure~\ref{fig:qwen_speedup}c, we additionally compare a datastore populated with the model’s own generated answers to one built from the original dataset answers, which follow a different distribution. We observe up to a 35\% increase in correct token predictions when using model-generated data.

The larger speedups observed for non-English languages are encouraging, and we observe the same pattern on the other tested models as well. With current tokenizers, equivalent content often requires fewer tokens in English than in many other languages, making English generation cheaper and faster \citep{petrov2023language,ahia2023do}. From this perspective, the larger gains of SSSD for other languages help mitigate tokenization-driven disparities and make generation latency more uniform across languages. We do not find clear evidence that morphological typology or other broad linguistic properties are major drivers of cross-lingual variation in speculation quality. We hypothesize that this variation is driven primarily by language-specific factors and their interaction with tokenization, although a more systematic investigation is left to future work.

\subsection{End-to-end evaluations with batching} \label{sec:main_results}

\begin{figure*}[!ht]
    \centering
    \begin{minipage}{0.32\textwidth}
        \centering
        \includegraphics[width=\textwidth]{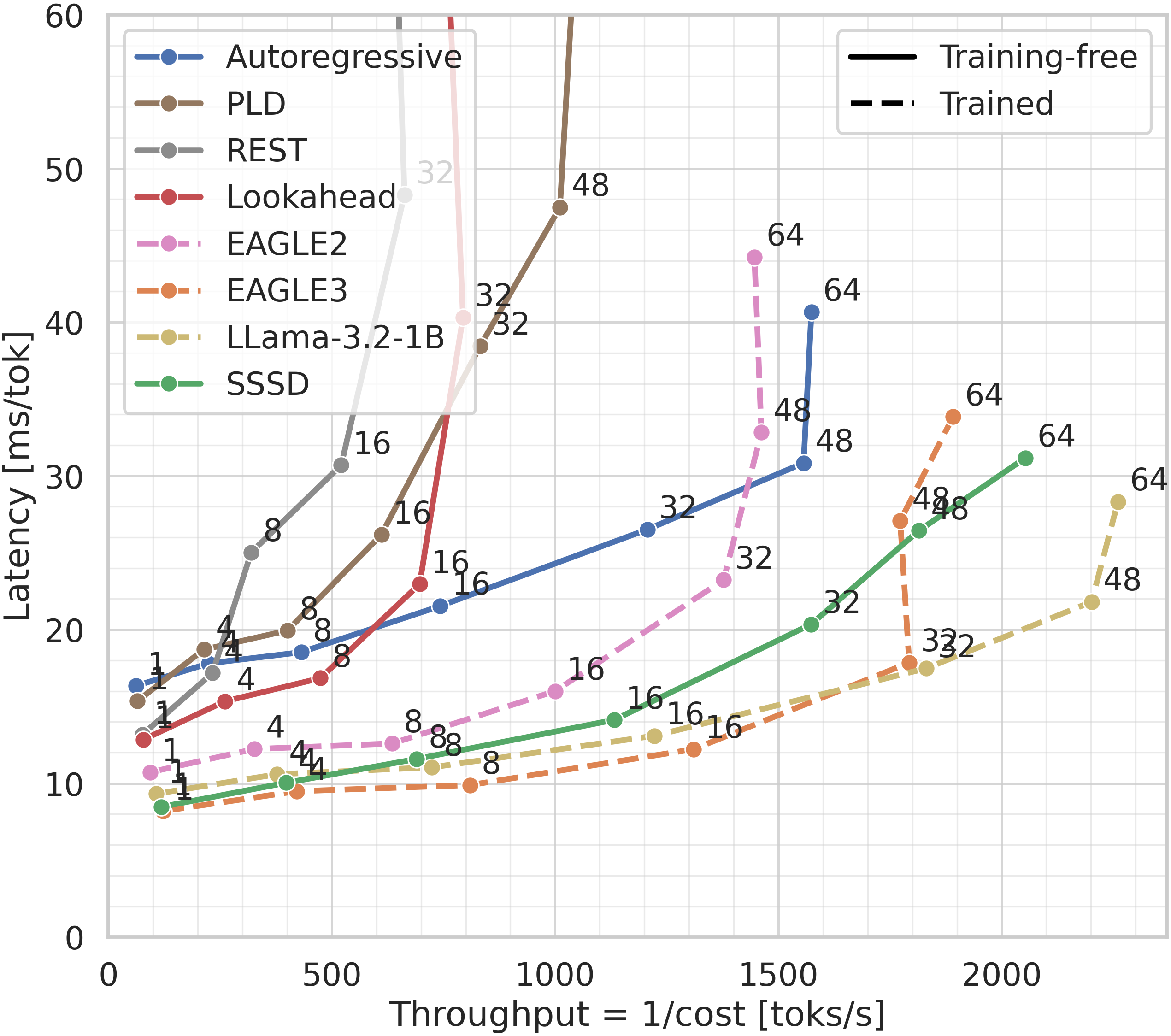}
        {\footnotesize (a) Llama-3.1-8B - MT-Bench}
    \end{minipage}
    \hfill
    \begin{minipage}{0.32\textwidth}
        \centering
        \includegraphics[width=\textwidth]{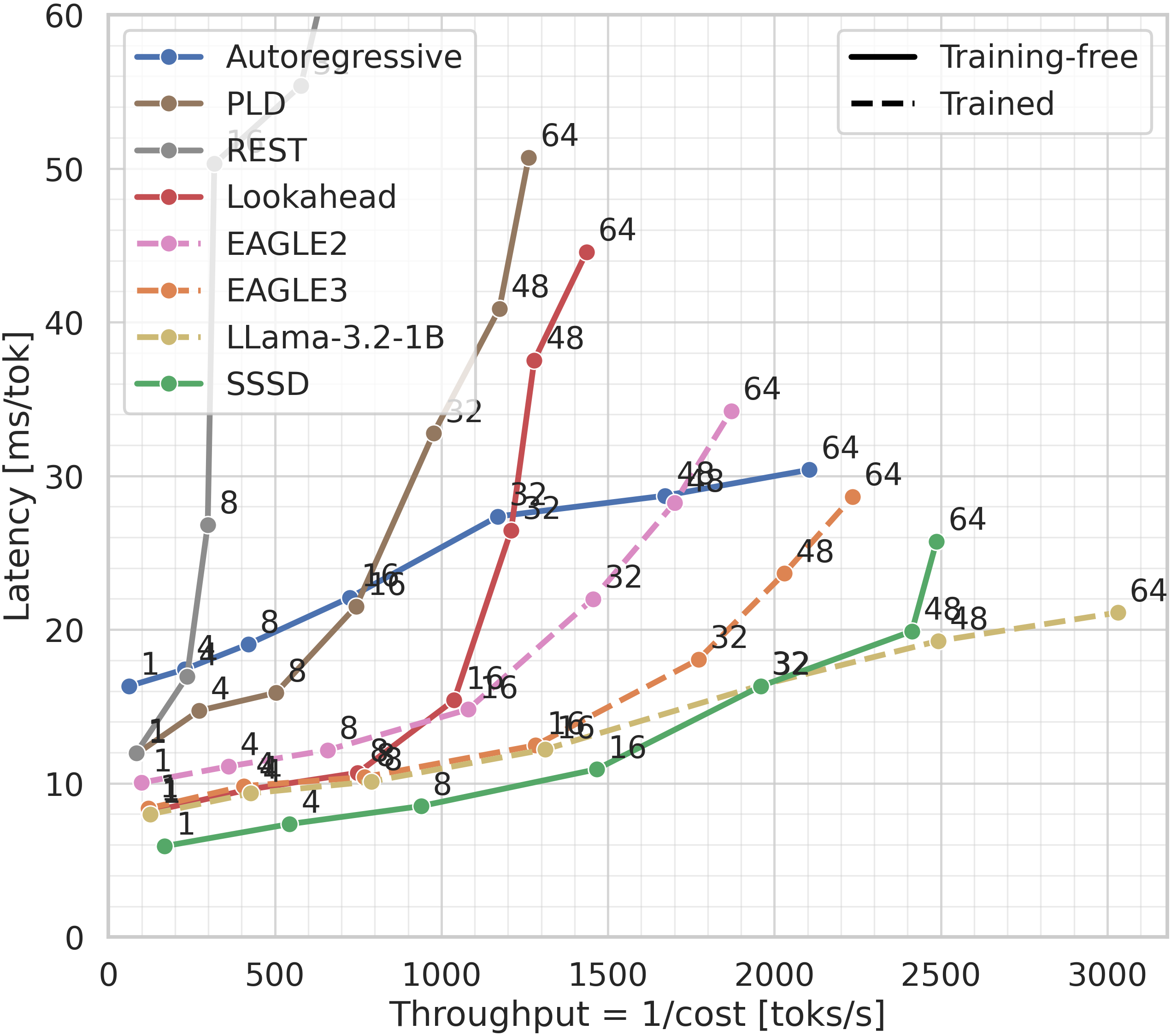}
        {\footnotesize (b) Llama-3.1-8B - MATH-500}
    \end{minipage}
    \hfill
    \begin{minipage}{0.32\textwidth}
        \centering
        \includegraphics[width=\textwidth]{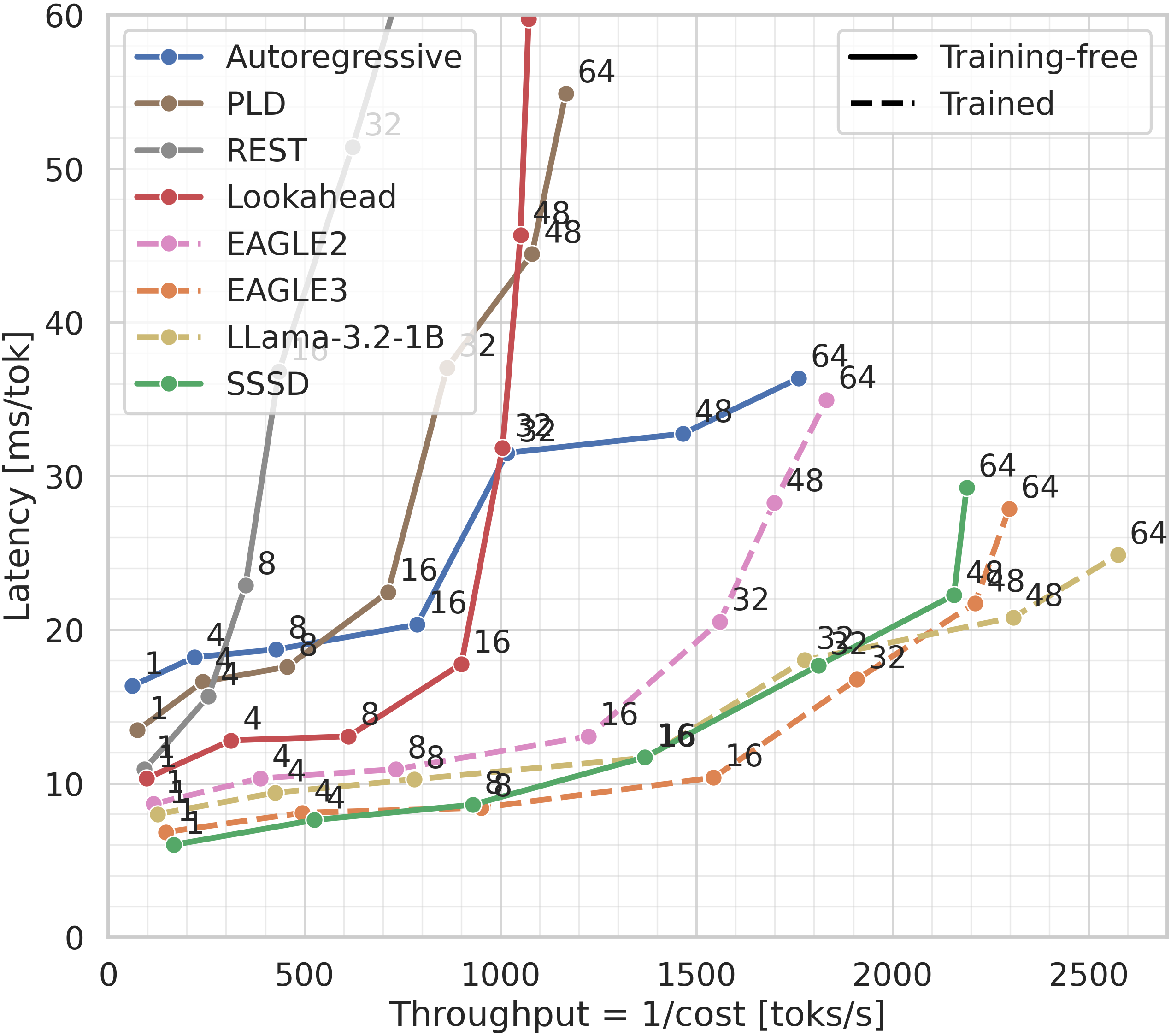}
        {\footnotesize (c) Llama-3.1-8B - HumanEval (Code)}
    \end{minipage}

    \vspace{0.5em} 

    \begin{minipage}{0.32\textwidth}
        \centering
        \includegraphics[width=\textwidth]{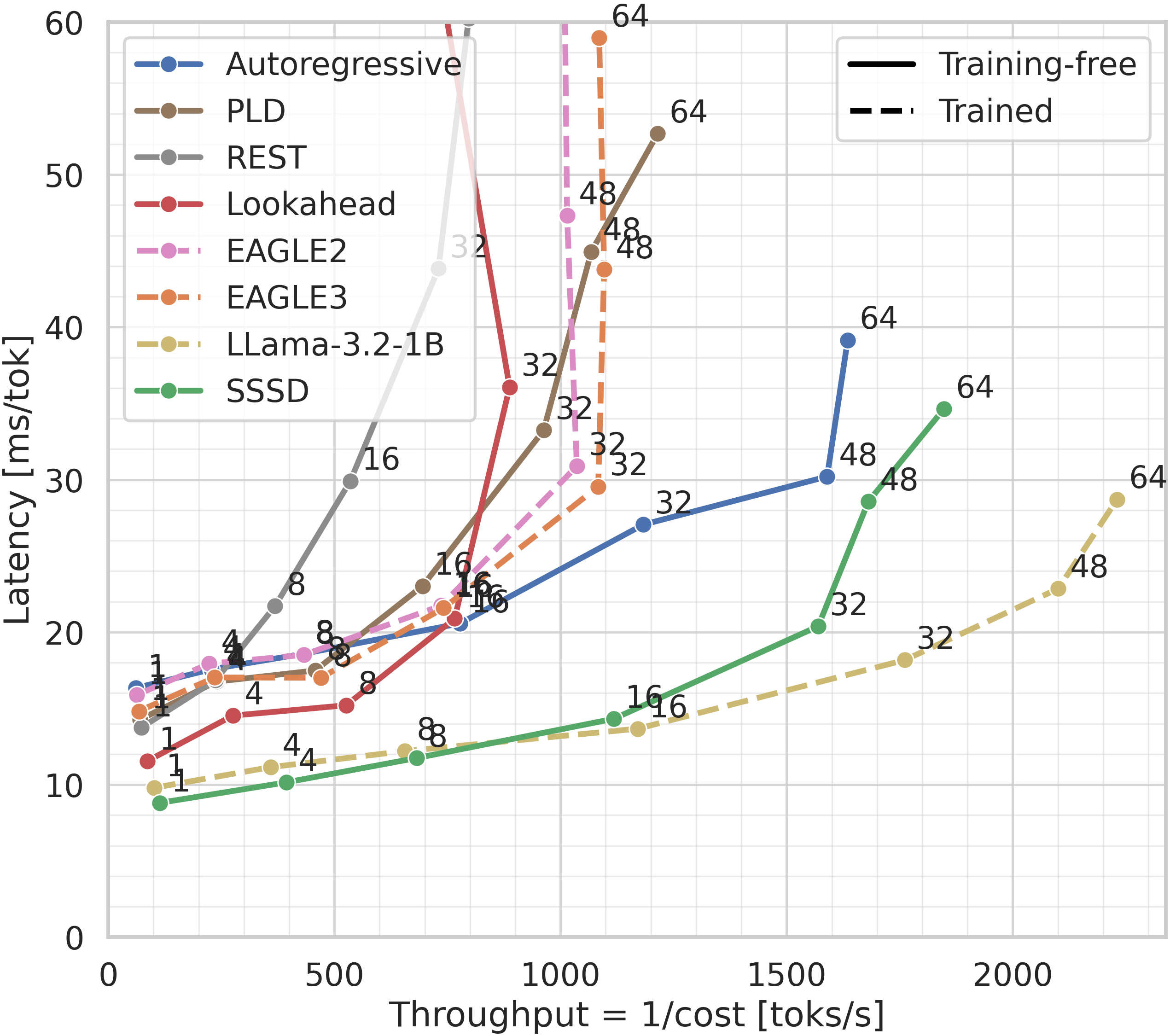}
        {\footnotesize (d) Llama-3.1-8B - MT-Bench German}
    \end{minipage}
    \hfill
    \begin{minipage}{0.32\textwidth}
        \centering
        \includegraphics[width=\textwidth]{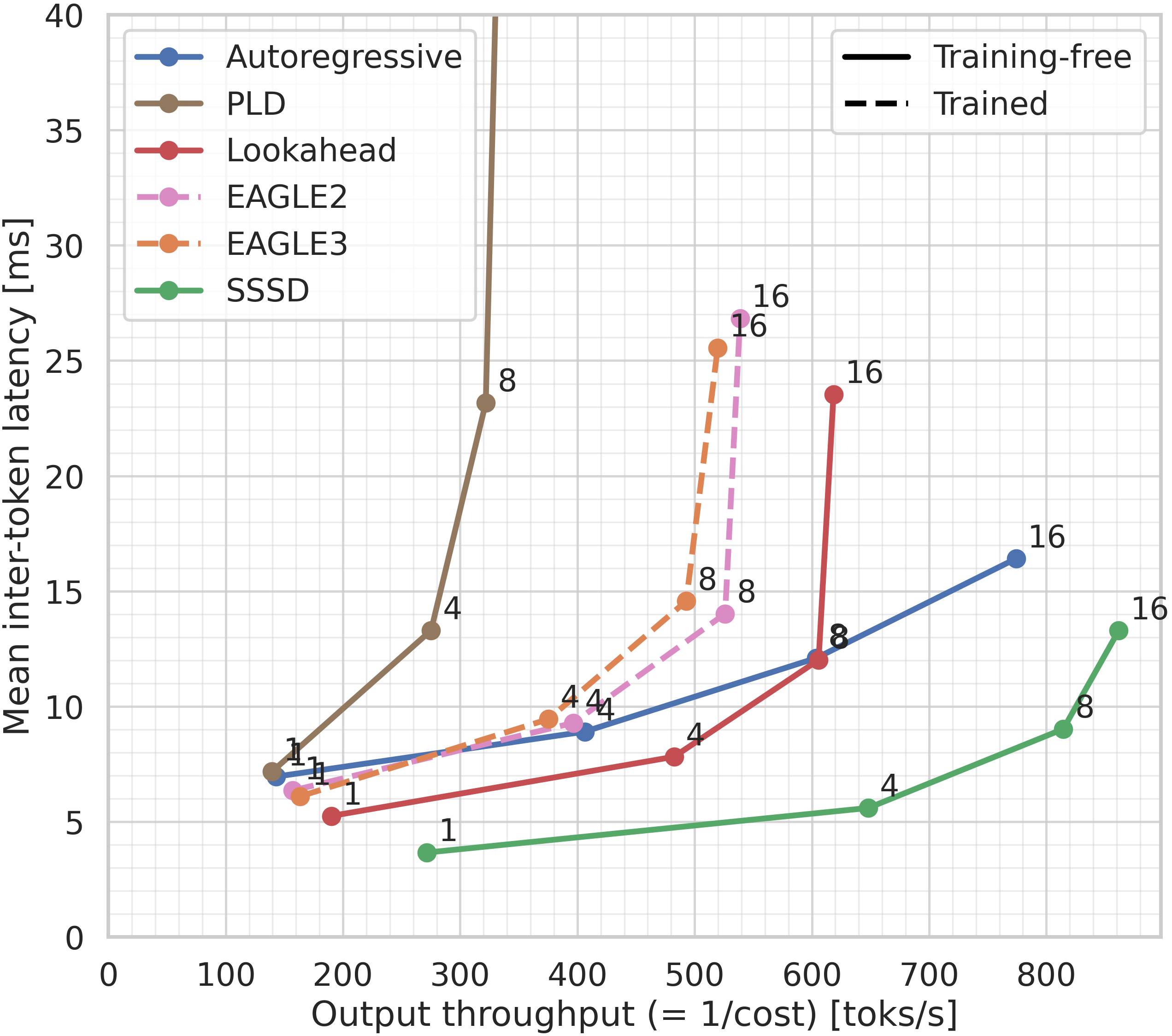}
        {\footnotesize (e) Llama-3.1-8B - SWE-Bench}
    \end{minipage}
    \hfill
    \begin{minipage}{0.32\textwidth}
        \centering
        \includegraphics[width=\textwidth]{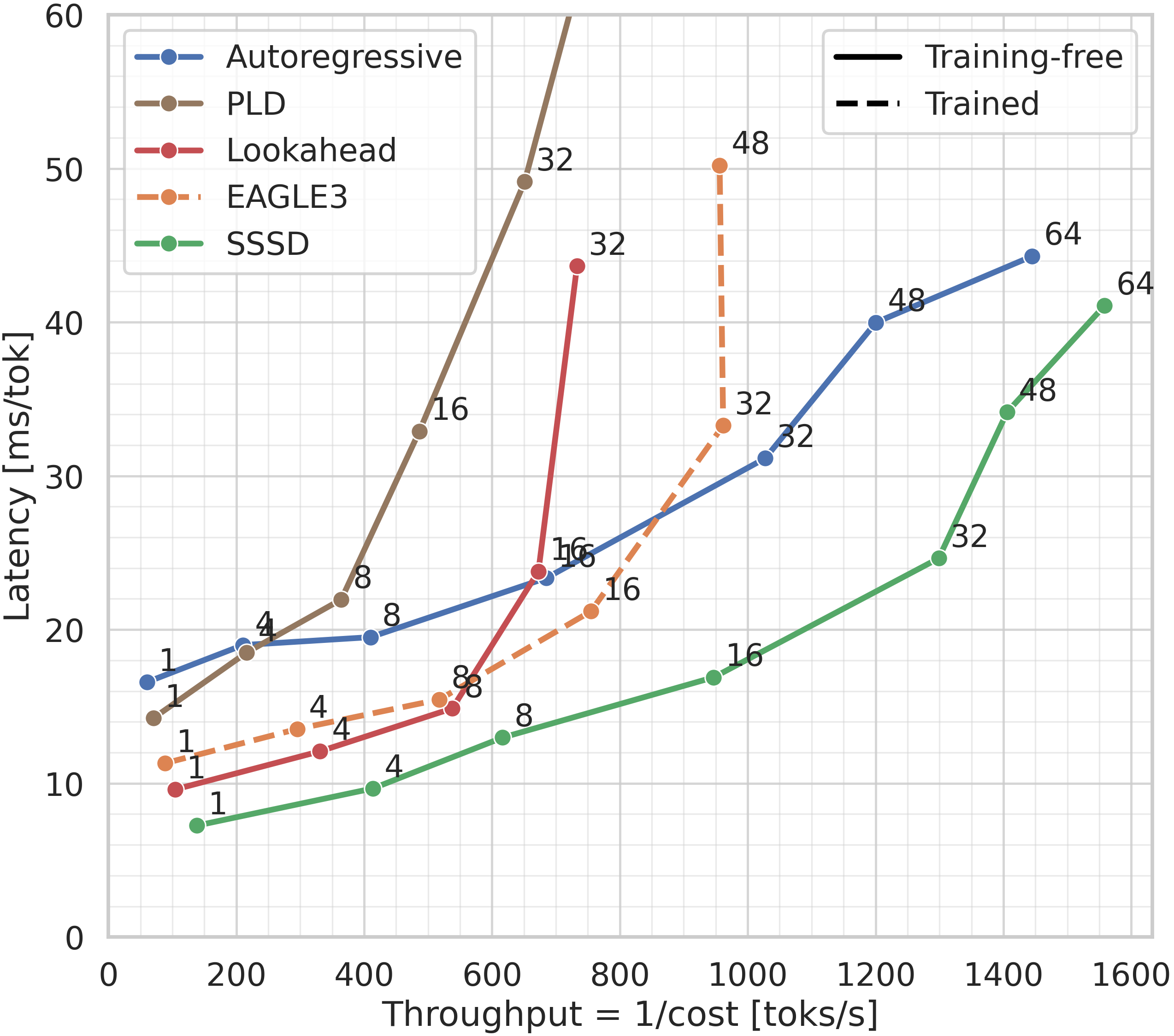}
        {\footnotesize (f) DS-R1-Dist-Llama-8B - MATH-500}
    \end{minipage}

    \caption{Evaluation of speculation methods on 8B models. (a--d) Llama-3.1-8B on a single GPU (same as in Figure~\ref{fig:qwen_speedup}) across multiple datasets. (e) Llama-3.1-8B in disaggregated prefill–decode mode on 4 GPUs (1 for decoding, 3 for prefill), each with 80 GB of VRAM, 989.5 TFLOPS, and 3.35 TB/s memory bandwidth. (f) DeepSeek-R1-Distill-Llama-8B on MATH-500 using the same hardware as in (a), averaged over five runs.}
    \label{fig:results_8b}
\end{figure*}

For the experiments on Llama-3.1-8B, we use the same datastore for SSSD and REST, starting from the same data as in Section \ref{sec:acceptance_length}. To test cross-lingual adaptation, we add a small amount of German data (not generated by the model), consisting of translations of ShareGPT \citep{freedomintelligence_multilingual} and Synthia \citep{Synthia_2021} samples. We also add a subset of The Stack \citep{Kocetkov2022TheStack} to improve performance on coding benchmarks. It is worth recalling that if all data were generated directly from the target model, speculation accuracy would be higher. The resulting datastore has a total size of 14 GB. We use the same datastore for all evaluated datasets, highlighting SSSD’s ability to retrieve the relevant information for each task when the datastore contains heterogeneous text.
For Lookahead, we do not perform cache warm-up, as it is not supported in SGLang and does not consistently yield performance improvements (see Figure~\ref{fig:method_comparisons}c).

Results are shown in Figure~\ref{fig:results_8b}, while results for Llama-3.3-70B are reported in Appendix~\ref{app:large_model}. We present them in a latency–throughput plot to illustrate the relationship between these two key metrics across batch sizes. We omit time-to-first-token (TTFT), as it is mostly determined by the prefill stage and is therefore not affected by SSSD. In contrast to model-based methods that require constructing (and potentially transmitting) a draft KV-cache, SSSD builds the input tree in a background thread during prefill and requires no additional KV-cache.

When running on a single device, we operate in offline mode (all prompts are available at the start). We measure the output-token throughput ($\text{output tokens} / \text{total time}$) and compute the average latency ($\text{batch size} / \text{throughput}$). The total time includes the prefill stage, which SD does not accelerate. Across all datasets and batch sizes, we find that SSSD consistently outperforms every n-gram method and EAGLE-2. It also surpasses EAGLE-3 on MATH-500 \citep{hendrycks2021measuring} and on German-language questions. For general short-context conversational queries (MT-Bench~\citep{mtbench}) and short-context coding tasks (HumanEval~\citep{humaneval}), EAGLE-3 maintains a slight advantage. Nonetheless, the performance gap remains small despite the simplicity of SSSD compared to EAGLE.

We also report standard speculative decoding results using Llama-3.2-1B as the draft model. Trained under the same training regime as the target model on 9 trillion tokens with 916k GPU hours \cite{llama3.2}, it closely matches the target’s output distribution and achieves the highest acceptance rates. While informative, this setting is impractical when similarly aligned draft models are unavailable; nevertheless, SSSD achieves comparable end-to-end speedups.

\subsubsection{Long context}

Long-context generation is increasingly important for LLM applications such as agentic AI and retrieval-augmented generation (RAG) \citep{rag2020}. In this regime, SSSD has two key advantages over model-based speculative decoding methods.
First, SSSD eliminates the need for a KV cache. Although draft models and heads are smaller than the target model, their computational cost still grows with context length. In contrast, SSSD maintains an approximately constant drafting cost as the context length increases.
Second, SSSD’s speculation quality remains stable independently of input size. This is because the context used for n-gram lookups consists of only a few tokens, whereas speculative models require long-context training to avoid accuracy degradation.

Speculative decoding is particularly beneficial in the long-context regime due to the memory-bandwidth bottleneck of attention layers (Section~\ref{main:roofline_modeling}). While current SGLang overheads limit achievable gains in practice, SSSD still outperforms alternative methods on PG-19 \citep{pg19} (Table~\ref{tab:long_context}). On agentic workloads (Figures~\ref{fig:results_8b}e and~\ref{fig:results_70b}f), SSSD achieves up to 1.9× throughput and latency improvements over autoregressive decoding and 1.66× over EAGLE-3.

\begin{table}[ht]
    \centering
    \footnotesize
    \setlength{\tabcolsep}{3pt}
    \begin{tabularx}{\columnwidth}{lYY}
        \toprule
        & Llama-3.1-8B & Llama-3.3-70B \\ \midrule
        Lookahead  & 1.08×          & 1.15×          \\
        EAGLE-3 & 0.80×          & 1.09×          \\
        SSSD   & \textbf{1.23×} & \textbf{1.26×}        \\
        \bottomrule
    \end{tabularx}
    \caption{Speedup over autoregressive decoding in long-context summarization on PG-19 (up to 40k tokens), using PD disaggregation.}
    \label{tab:long_context}
\end{table}

\subsection{Reasoning and speculative sampling} \label{sec:reasoning}

Another important recent application of LLMs is reasoning models: in order to answer more complex questions the model is prompted~\citep{wei2022cot} or trained \citep{deepseekR1} to produce a reasoning trace before giving a final answer. This substantially increases generation length, making speculative decoding particularly important.

We evaluate one of the distilled DeepSeek-R1 models \citep{deepseekR1} on a subset of medium- and hard-difficulty questions from MATH-500, allowing generation of up to 4096 tokens. The datastore is constructed using publicly available reasoning traces from larger models within the same family, resulting in an 18 GB datastore. We follow the recommended generation settings (temperature = 0.6, top-p = 0.95), demonstrating that our method extends beyond greedy decoding. We adopt a simplified speculative sampling procedure \citep{leviathan2023}, as implemented in SGLang, which preserves the target distribution without requiring draft probability estimates.

Results are shown in Figure~\ref{fig:results_8b}f: SSSD achieves a 2.29× speedup over autoregressive decoding at batch size 1 and a 1.32× speedup over the strongest alternative method. Notably, it is the only method that achieves speedups at large batch sizes.

\section{Conclusion}

This work argues for a re-centering of speculative decoding around deployability rather than speculation quality alone. SSSD demonstrates that strong and reliable acceleration can be achieved without auxiliary training, specialized draft models, or task-specific tuning, showing state-of-the-art performance under distribution shift, multilingual inputs, and long-context workloads. By framing inference acceleration as a joint algorithm–system problem, SSSD highlights a path toward simpler and more scalable decoding strategies that better align with the operational constraints of modern LLM serving.

\section*{Limitations}

In Mixture of Experts (MoE) models \citep{jiang2024mixtralexperts}, increasing speculation length activates more experts and thus requires loading more of them from memory, unless all of them are already loaded due to a large batch size. In this case, SSSD is affected similarly to other speculative decoding methods, yielding smaller gains for MoE models. Speedups are also limited for models using MLA attention \citep{deepseekv2}, which is typically run using the efficient, but compute-bound, FlashMLA kernels, which are less compatible for speculative decoding than standard MHA/GQA. In such scenarios, methods that prioritize higher speculation accuracy---at higher computational cost---may be more effective, as achievable speculation depth is inherently constrained.

Additionally, the drafting phase of SSSD depends on the CPU performance and requires some host memory capacity, which may limit its applicability in some deployment environments.

\section*{Ethical considerations}

Speculative decoding methods can introduce uneven performance across languages, tasks, and user populations. Prior work has shown that training-based draft models often generalize poorly beyond the domains and languages represented in their training data, leading to substantially lower speculation accuracy--and consequently lower inference speedups--for underrepresented languages and tasks. In production systems, such disparities translate into higher latency and cost for some users, effectively creating unequal access to efficient LLM services.

SSSD exhibits substantially more uniform speedups across languages, including lower-resource languages, without requiring language-specific training, tuning, or maintenance. While this does not address broader issues of linguistic coverage or bias in the underlying language model, it reduces a specific source of disparity introduced by existing inference acceleration techniques.

Finally, as with other retrieval-based methods, using prior model outputs as a source of speculative candidates raises privacy considerations. In practice, SSSD operates on transient data already maintained by many serving systems, preserves the target model’s output distribution, and does not introduce additional information leakage in generated outputs.


\bibliography{custom}

\appendix

\section{Algorithms}
\label{sec:appendix_algo}

This section describes the main algorithms used for candidate retrieval.
The algorithms presented here are not exact replicas of the implementation but are logically equivalent; minor differences are introduced to improve clarity. We omit pseudocode for constructing the underlying data structures and denote the cardinality of data structures (e.g., the total number of nodes in a tree or the number of elements in a list) by $|\cdot|$.

\subsection{Input tree}

To retrieve candidates from the prompt and self-output, we maintain a trie data structure whose nodes correspond to token prefixes. When a sequence of tokens is inserted into the trie, the count of every node along the corresponding path is incremented. Consequently, each node stores the number of times its associated prefix has been observed.

Based on these counts, each node is associated with probability values conditioned on different prefixes, denoted as node.\textit{prob}. For a prefix represented by a node $u$ and a longer sequence represented by a descendant node $v$, this probability is defined as
\[
    \text{node}.\textit{prob} = \frac{\text{count}(v)}{\text{count}(u)}.
\]
This quantity represents the empirical probability that the prefix corresponding to $u$ is extended to the sequence corresponding to $v$; the extension need not be immediate.

When retrieving continuations for multiple prefixes of varying lengths from the input trie, we return references to the nodes corresponding to the ends of those prefixes. Due to the recursive structure of the trie, each such node implicitly defines a complete candidate tree, namely the subtree rooted at that node.

The input trie, together with its node counts, is constructed incrementally and reused across forward passes for the same prompt. At each iteration, the trees of candidates from the input are simply subtrees of the input trie. In contrast, the datastore tree is recomputed at each decoding step based on the candidate continuations identified by Algorithm~\ref{alg:get_cands}.

\begin{algorithm}[!t]
\caption{Retrieving candidates from the input tree for a given prefix. The tree typically has a maximum depth below 10.}
\label{alg:input_retrieval}
\begin{algorithmic}
\Function{MatchInput}{prefix}
    \State {\bfseries Input:} T: Tree of prompt + current output
    \State results $\gets$ [\,]
    \State iter $\gets 0$
    \While{iter $< |$prefix$|$}
        \State p $\gets$ prefix[iter..] \Comment{sub-prefix}
        \State node $\gets$ T.\Call{Root}{}
        \State $i \gets$ 0
        \While{i $<|$p$|$ \textbf{and} p[i]$\in$ node.\textit{children}}
            \State node $\gets$ node.\textit{children}[p[i]]
            \State $i \gets i+1$
        \EndWhile
        \If{i $= |$p$|$} \Comment{sub-prefix found}
            \State results.\Call{Append}{$\langle$node, $|$p$|\rangle$}
        \EndIf
        \State iter $\gets$ iter$ + 1$
    \EndWhile
    \State \Return results
\EndFunction
\end{algorithmic}
\end{algorithm}

\begin{algorithm}[!t]
\caption{Retrieving candidate continuations for a given prefix from a suffix array.}
\label{alg:get_cands}
\begin{algorithmic}
\Function{GetCands}{prefix, branch\_len}
    \State tree $\gets$ \Call{Tree}{root\_id$\gets$prefix[$|$prefix$|-1$]}
    \State iter $\gets 0$
    \While{iter $< |$prefix$| - 1 $ \textbf{and} $|$tree$| < 50$}
        \State prefix $\gets$ prefix[iter..]
        \State conts $\gets$ index.\Call{BinarySearch}{}(
        \State \hspace{8em} prefix, branch\_len)
        \State step $\gets \max\left(1,\left\lfloor \dfrac{\left|\text{conts}\right|}{100} \right\rfloor\right)$
        \State i $\gets$ 0
        \While{i $< |$conts$|$}
            \State cont $\gets$ conts[i]
            \State tree.\Call{Insert}{cont}
            \State i $\gets$ i + step
        \EndWhile
        \State iter $\gets$ iter$+ 1$
    \EndWhile
    \State \Return tree
\EndFunction
\end{algorithmic}
\end{algorithm}

\begin{algorithm}[!ht] 
   \caption{Merging candidate tokens coming from different sources. The method \protect\Call{InsertTok}{} on the draft tree takes a token and a parent node, adds the token as a child of the parent if it does not exist yet, and returns the child node.}
   \label{alg:fusing_sources}
\begin{algorithmic}
   \State {\bfseries Input:} prefix (vector of integers), dec\_len (max final number of candidates), branch\_len (max depth of each draft)
   \State {\bfseries Output:} draft\_tree (the merged tokens)
   \State input\_trees $\gets$ input\_cache.\Call{MatchInput}{}(
   \State \hspace{9em} prefix)
   \State datastore\_tree $\gets$ datastore.\Call{GetCands}{}(
   \State \hspace{9em} prefix, branch\_len)
   \State pq \(\leftarrow \emptyset\) \Comment{priority queue}
   \State draft\_tree $\gets$ \Call{Tree}{root\_id$\gets$prefix[$|$prefix$|-1$]}
   \State all\_trees $\gets$ [$\langle$datastore\_tree, 4$\rangle$] $\mathbin{\|}$ input\_trees
   \For{\textbf{each} $\langle$tree, match$\rangle$ $\in$ all\_trees}
      \If{tree \(\neq \emptyset\)}
         \State node $\gets$ tree.\Call{Root}{}
         \For{\textbf{each} child $\in$ node.\textit{children}}
             \If{tree = datastore\_tree}
                \State discount $\gets$ 1
             \Else
                \State discount $\gets$ 0.6
             \EndIf
             \State child\_disc $\gets$ 0.6 + 0.1 $\cdot$ match
             \State prob $\gets$ node.\textit{prob} $\cdot$ discount
             \State pq.\Call{Push}{}(prob,  $\langle$\textit{node}$\gets$child,
             \State \hspace{3em} \textit{child\_disc}$\gets$child\_disc,
             \State \hspace{3em} \textit{parent}$\gets$draft\_tree.\Call{Root}{}$\rangle$)
         \EndFor
      \EndIf
   \EndFor
   \While{pq $\neq \emptyset$ \textbf{and} $|$draft\_tree$| <$ dec\_len}
      \State prob, el $\gets$ pq.\Call{Pop}{\null}
      \State new\_candidate $\gets$ draft\_tree.\Call{InsertTok}{}(
      \State \hspace{4em} el.\textit{node.token\_id}, el.\textit{parent})
      \For{\textbf{each} child $\in$ el.\textit{node.children}}
         \State p $\gets$ child.\textit{prob} $\cdot$ prob $\cdot$ el.\textit{child\_disc}
         \State pq.\Call{Push}{}(p, $\langle$\textit{node}$\gets$child,
         \State \hspace{4em} \textit{child\_disc}$\gets$el.\textit{child\_disc}
         \State \hspace{4em} \textit{parent}$\gets$new\_candidate$\rangle$)
      \EndFor
   \EndWhile
   \State \textbf{return} draft\_tree
\end{algorithmic}
\end{algorithm}

\subsection{Datastore}

As explained in Section \ref{main:datastore_management}, the datastore consists of multiple suffix arrays (capped at most at 8). The suffix arrays follow a standard implementation \citep{suffix_array1990}: given a prefix, its possible continuations are obtained by performing two binary searches to locate the first and last lexicographic occurrences of the prefix.
Let $P$ be the prefix length and $N$ the number of tokens in a sub-index. For each sub-index, the cost of the two binary searches is $O(P \log N)$. The total search cost increases linearly with the number of sub-indices, which is capped and small.

The remaining work (merging continuations and selecting candidates) is independent of datastore size because the number of sampled continuations is fixed. In practice, larger datastores often reduce retries (e.g., fewer fallbacks to $(P = 3)$), partially offsetting the cost increase.

The construction of the tree of candidates coming from the datastore is straightforward: inserting a selected continuation consists of adding any missing nodes along its path (each initialized with a count of 1) and incrementing the counts of all nodes that already exist along that path.

\subsection{Merging input and datastore candidates}

After assigning weights to the candidate sources as described in Section \ref{main:fusion}, we select the final draft tokens using a priority queue, as detailed in Algorithm \ref{alg:fusing_sources}. The cost of this step is independent of the input and datastore size.

\section{Impact of the key variables of the decoding phase}
\label{app:roofline}

\begin{table*}[t]
\caption{Summary of FLOPs and memory IO for a speculative forward pass. We use $b$ for batch size, $s_q$ for speculation length, $s_{kv}$ for KV-cache length, $n_q$ and $n_{kv}$ for query and key/value heads, $g=n_q/n_{kv}$, $d$ for per-head dimension, $h$ for the model's hidden dimension ($h=n_q d$), $h_{mlp}$ for the MLP intermediate dimension, and $p$ for tensor parallelism. We assume fused scaled-dot-product attention~\citep{dao2022flashattention, dao2024flashattention2}. Input, local KV-cache, and mask tensors have shapes $X$: $(b, s_q, h)$, $K_{\text{cache}}, V_{\text{cache}}$: $(b, n_{kv}/p, s_{kv}, d)$, and $M$: $(b, s_q, s_{kv}+s_q)$. Per-rank local weight shapes are $W_q$: $(h, h/p)$, $W_k, W_v$: $(h, h/(gp))$, $W_o$: $(h/p, h)$, $W_1$: $(h, 2h_{mlp}/p)$, and $W_2$: $(h_{mlp}/p, h)$, with $W_1$ denoting the concatenated SwiGLU input projections. The remaining tensor shapes are $Q$: $(b, n_q/p, s_q, d)$, $K_{\text{new}}, V_{\text{new}}$: $(b, n_{kv}/p, s_q, d)$, $K, V$: $(b, n_{kv}/p, s_{kv}+s_q, d)$, $A_{out}$: $(b, n_q/p, s_q, d)$, $Y, Z$: $(b, s_q, h)$, and $H$: $(b, s_q, h_{mlp}/p)$. Memory IO is in bytes, with 2 bytes per weight, activation, and KV-cache element, and 1 byte per mask element. We assume identical $s_{kv}$ across sequences, count only GEMM FLOPs in attention, and exclude tensor-parallel communication.}
\centering
\fontsize{8pt}{9.5pt}\selectfont
\setlength{\tabcolsep}{4pt}
\renewcommand{\arraystretch}{1.4}

\begin{tabular*}{\textwidth}{@{\extracolsep{\fill}} c c c c c c @{}}
\toprule
\textbf{Op} &
\textbf{FLOPs} &
\textbf{Read (bytes)} &
\textbf{Write (bytes)} &
\textbf{FLOPs:IO} &
\textbf{FLOPs:IO appr.} \\
\midrule

$Q = XW_q$ &
\dfraccell{\frac{2 b s_q h^{2}}{p}} &
\dfraccell{2 \left(b s_q h + \frac{h^{2}}{p} \right)} &
\dfraccell{2 \frac{b s_q h}{p}} &
\dfraccell{\dfrac{1}{\dfrac{p+1}{h} + \dfrac{1}{b s_q}}} &
\makecell{\dfraccell{\mathbf{b s_q}}\\if $h \gg (p+1)b s_q$} \\

\tallrow $K_{\text{new}} = XW_k$ &
\dfraccell{\frac{2 b s_q h^{2}}{g p}} &
\dfraccell{2\left(b s_q h + \frac{h^{2}}{g p}\right)} &
\dfraccell{2\frac{b s_q h}{g p}} &
\dfraccell{\dfrac{1}{\dfrac{g p+1}{h} + \dfrac{1}{b s_q}}} &
\makecell{\dfraccell{\mathbf{b s_q}}\\if $h \gg (gp+1) b s_q$} \\

$V_{\text{new}} = XW_v$ &
same as $K_{\text{new}}$ &
same &
same &
same &
same \\

\tallrow \makecell{$A_{out}=\operatorname{softmax}$\\$\!\Bigl(\tfrac{Q K^{\mathsf T}}{\sqrt{d}} + M \Bigr)\, V$} &
\dfraccell{\frac{4 b s_q (s_{kv}+s_q) h}{p}} &
\makecell{%
\dfraccell{2\left(\frac{b h}{g p}\bigl(2 (s_{kv}+s_q)+g s_q\bigr)\right)}\\
\dfraccell{+\,b s_q (s_{kv}+s_q)}%
} &
\dfraccell{2\frac{b s_q h}{p}} &
\dfraccell{\dfrac{1}{\dfrac{1}{g s_q} + \dfrac{1}{s_{kv}+s_q} + \dfrac{p}{4 h}}} &
\makecell{%
\dfraccell{\mathbf{g s_q}}\\if $h \gg p g s_q/4$,\\$s_{kv} + s_q \gg g s_q$} \\

\tallrow $Y = A_{out} W_o$ &
\dfraccell{\frac{2 b s_q h^{2}}{p}} &
\dfraccell{2\left(\frac{b s_q h}{p} + \frac{h^{2}}{p}\right)} &
\dfraccell{2 b s_q h} &
\dfraccell{\dfrac{1}{\dfrac{p+1}{h} + \dfrac{1}{b s_q}}} &
\makecell{\dfraccell{\mathbf{b s_q}}\\if $h \gg (p+1)b s_q$} \\

\tallrow \makecell{$H =$\\$\mathrm{SwiGLU}(Y W_{1})$} &
\dfraccell{\frac{4 b s_q h h_{mlp}}{p}} &
\dfraccell{2\left(b s_q h + 2\,\frac{h h_{mlp}}{p}\right)} &
\dfraccell{2\frac{b s_q h_{mlp}}{p}} &
\dfraccell{\dfrac{1}{\dfrac{1}{2 h} + \dfrac{p}{2 h_{mlp}} + \dfrac{1}{b s_q}}} &
\makecell{\dfraccell{\mathbf{b s_q}}\\if $h \gg b s_q / 2$, \\$h_{mlp} \gg p b s_q/2$} \\

\tallrow $Z = H\,W_{2}$ &
\dfraccell{\frac{2 b s_q h h_{mlp}}{p}} &
\dfraccell{2\left(\frac{b s_q h_{mlp}}{p} + \frac{h h_{mlp}}{p}\right)} &
\dfraccell{2 b s_q h} &
\dfraccell{\dfrac{1}{\dfrac{1}{h} + \dfrac{p}{h_{mlp}} + \dfrac{1}{b s_q}}} &
\makecell{\dfraccell{\mathbf{b s_q}}\\if $h\gg b s_q$, \\ $h_{mlp} \gg p b s_q$} \\
\bottomrule
\end{tabular*}
\label{tab:formulas}
\end{table*}

This section extends the analysis on the interactions between speculative decoding and hardware utilization, in particular under batching. Larger batch sizes lead to resource competition between batching and speculation, limiting SD speedup. Some studies \citep{su2023synergyspeculativedecodingbatching,zhong2024propd} observe a linear relationship between forward pass overhead and speculation length, with steeper slopes at higher batch sizes. However, this empirical trend doesn’t align with theoretical hardware predictions, likely due to unoptimized operator implementations. Sequoia \citep{chen2024sequoia} shows empirically that the cost of the forward pass remains almost constant up to $s_q < N$, where $N$ depends on the hardware and the model. Understanding these interactions in detail is crucial for deploying SD in an inference system: it enables quantifying the potential gains of SD and it simplifies determining optimal speculation parameters at deployment, relying solely on hardware features.

Table~\ref{tab:formulas} shows the compute and memory costs of the main operations of an LLM, following \citet{lequn2023blog}. The equations are based on the Llama family \citep{grattafiori2024llama3herdmodels}, but most dense models have similar architectures. We integrate the speculation length ($s_q$) into FLOP and memory usage formulas for different transformer model components. When $s_q = 1$, the forward pass matches standard autoregressive decoding.

Increasing the batch size $b$ nearly proportionally increases the compute-to-memory access ratio for all terms except FlashAttention. Since memory accesses for matrix multiplications only increases slightly with $b$, the batch size can be scaled up with minimal extra costs until reaching the compute-bound region of the roofline model \citep{roofline2009}. Similarly, increasing the speculation length $s_q$ almost proportionally increases the compute-to-memory access ratio for all matrix multiplications, including FlashAttention. Here as well, memory accesses only slightly rise with $s_q$, so this term can be increased with negligible cost until reaching the compute-bound threshold.
Most importantly, for all matrix multiplications other than FlashAttention, the batch size and the speculation length have equivalent effects, always appearing as $b \times s_q$. Therefore, the theoretical free speculation budget can be computed from the hardware specifications by pushing $b \times s_q$ to the roofline model's compute-bound threshold. For instance, for the computation of the $Q$ matrix, assuming that there is no tensor parallelism, the speculation budget can be derived by solving

\begingroup
\begin{align*}
\dfrac{1}{(1/h + 1/b s_q)} = \frac{\text{Peak TFLOPS}}{\text{Memory Bandwidth (TB/s)}}.
\end{align*}
\endgroup

If we consider the Llama-3-8B \citep{grattafiori2024llama3herdmodels} model ($h = 4096$) and GPU with a measured computing power of 165\,TFLOP/s and a measured memory bandwidth of 0.95\,TB/s, the (almost) free speculation budget can be theoretically obtained until $b s_q \approx 174$. This limit ($s_q=22$ for $b=8$) appears as a discontinuity in the slope of the green and purple lines in Figure~\ref{fig:dissection}a, where the initial segment reflects memory-bandwidth limitations, and the latter segment is determined by the available FLOPS. In practice, the available operators often under-utilize the hardware, making the empirical gains smaller than what can be seen in Figure~\ref{fig:dissection}. This is because of the complexity involved in creating efficient kernels, in particular for matrix multiplications where one matrix is ``tall and skinny'' \citep{rivera2021}.

Finally, increasing the context length $s_{kv}$ has the main effect of increasing the FlashAttention's time. As a result, other matrix multiplications, unaffected by $s_{kv}$, have a lower overall weight in the forward pass (see Figure~\ref{fig:dissection}b). In this scenario, the cost of loading the KV-cache comprises the highest share, and is less impacted by a longer speculation length, unlike large matrix multiplications: the FLOPs:IO ratio ($\approx g s_q$), where $g$ denotes the group size in grouped-query attention (GQA), i.e., the number of query heads sharing a single key–value head, does not include the batch size, making the transition from memory-bandwidth bound to compute bound happen at much higher $s_q$ compared to the other operations if $g=1$. Otherwise, for $b < g$, the attention layer gets in the compute-bound region for shorter speculation lengths. Depending on the dominating cost between linear layers and attention, the ``free'' compute budget can be given by the FLOPs:IO ratio divided by the batch size (for short context), or by the group size (for long context). In Figure~\ref{fig:dissection}b, you can see that for very long context, although the orange region turns steeper already at $s_q = 22$, the overall increase in cost is negligible until the attention becomes compute bound. Note that in Figure~\ref{fig:dissection}b the batch size $b=8$ is larger than the group size $g=4$.

\begin{figure*}[ht]
    \centering
    \begin{minipage}{0.32\textwidth}
        \centering
        \includegraphics[width=\textwidth]{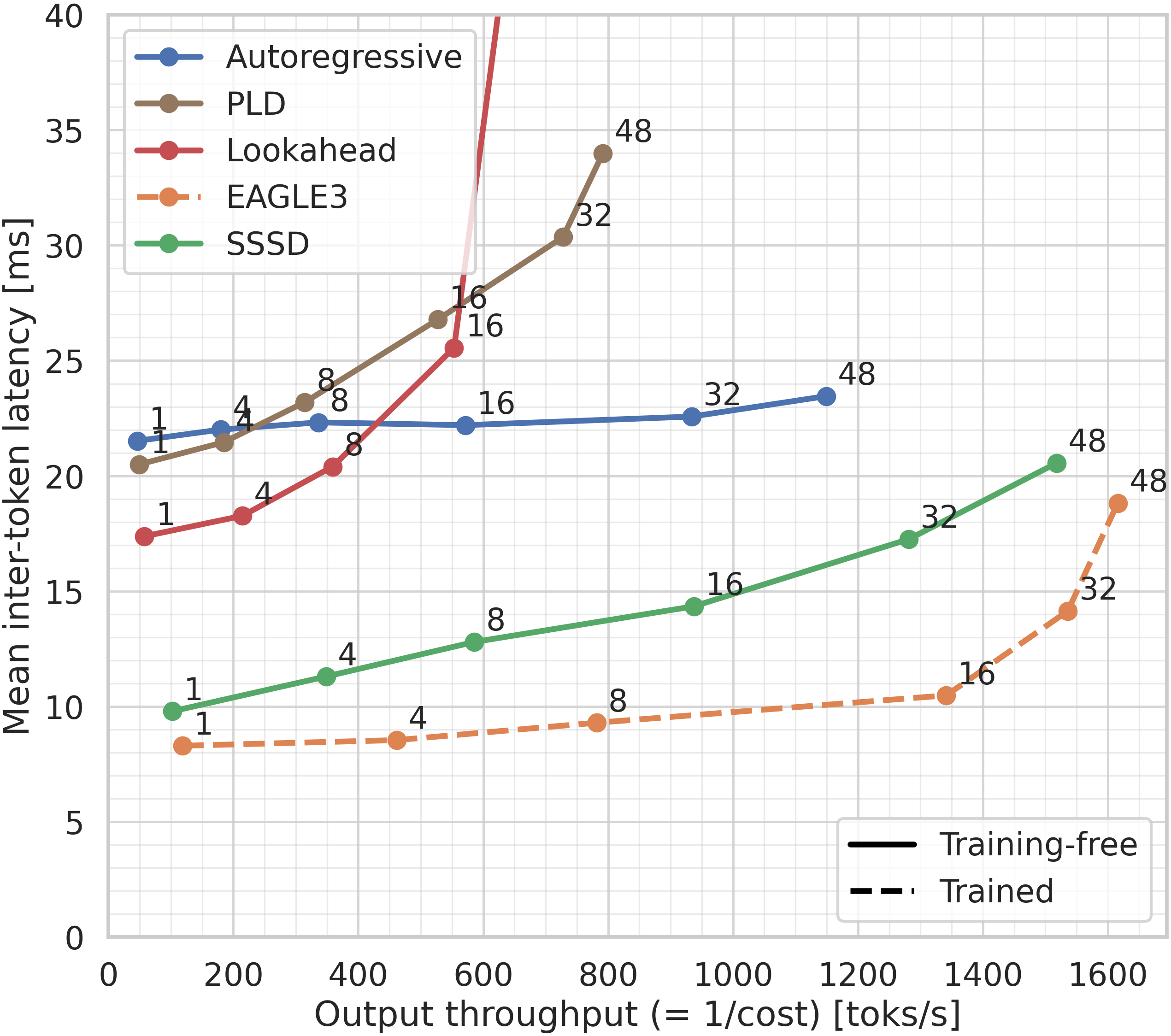}
        {\footnotesize (a) MT-Bench}
    \end{minipage}
    \hfill
    \begin{minipage}{0.32\textwidth}
        \centering
        \includegraphics[width=\textwidth]{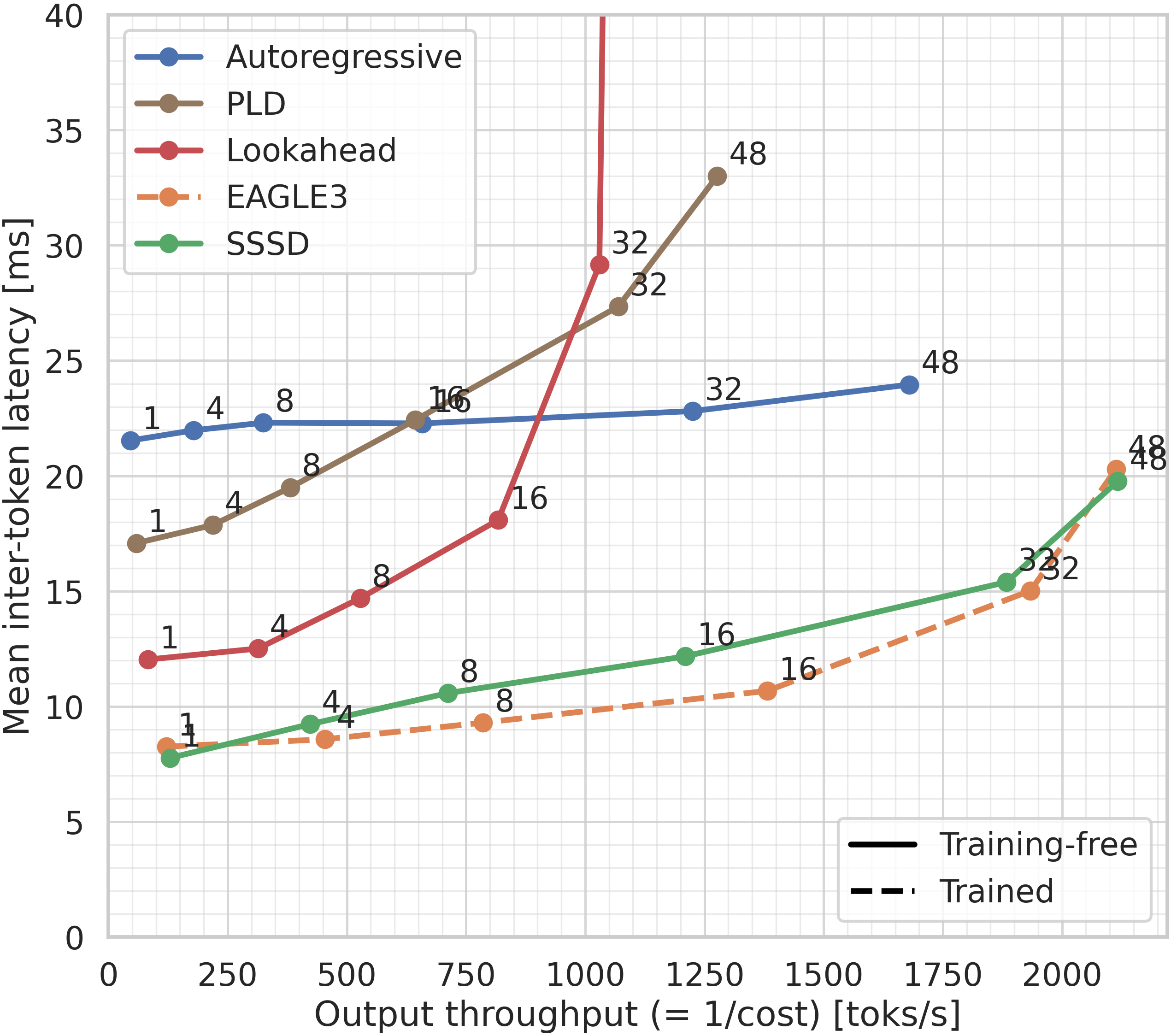}
        {\footnotesize (b) MATH-500}
    \end{minipage}
    \hfill
    \begin{minipage}{0.32\textwidth}
        \centering
        \includegraphics[width=\textwidth]{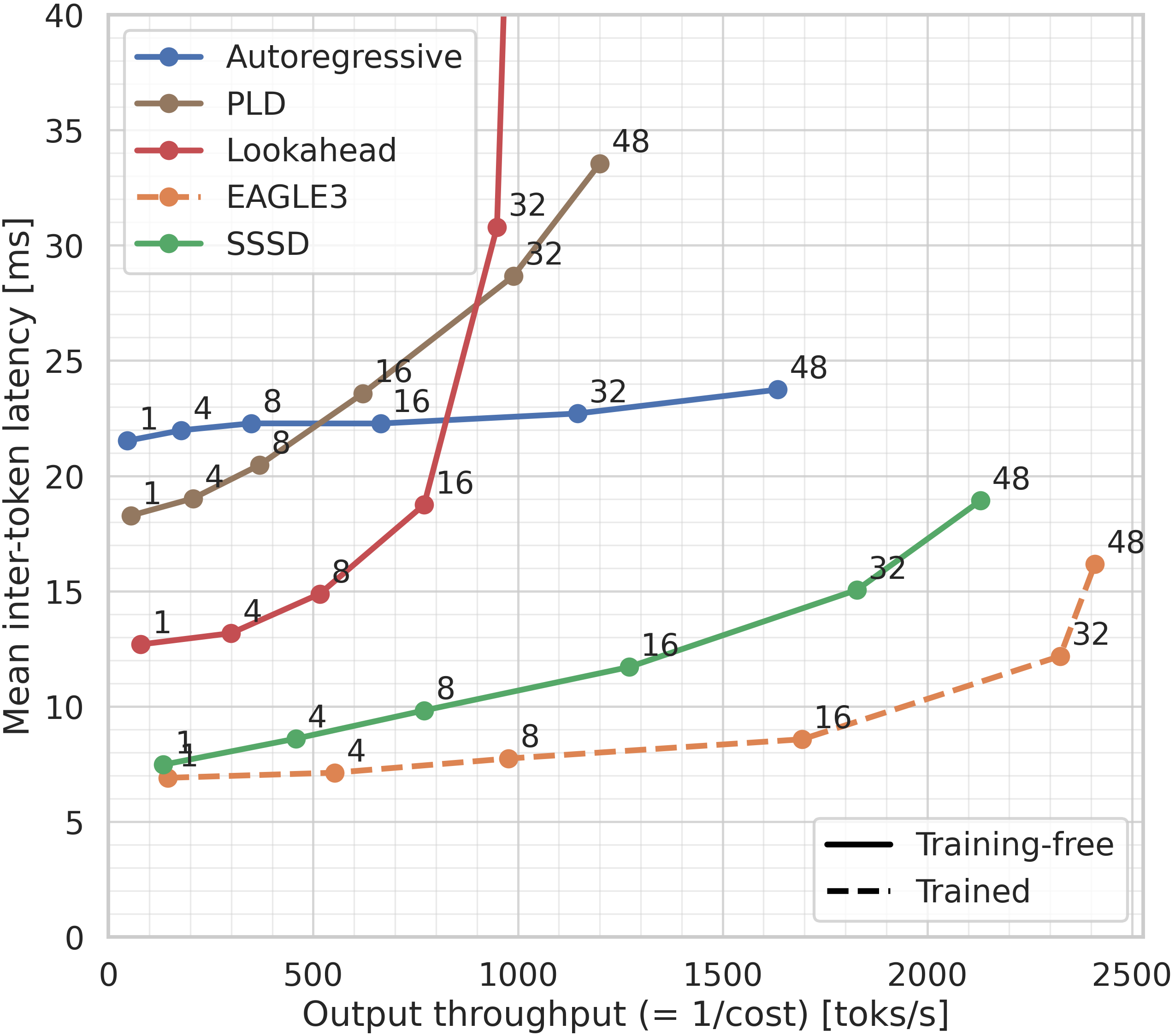}
        {\footnotesize (c) HumanEval (Code)}
    \end{minipage}

    \vspace{0.5em} 

    \begin{minipage}{0.32\textwidth}
        \centering
        \includegraphics[width=\textwidth]{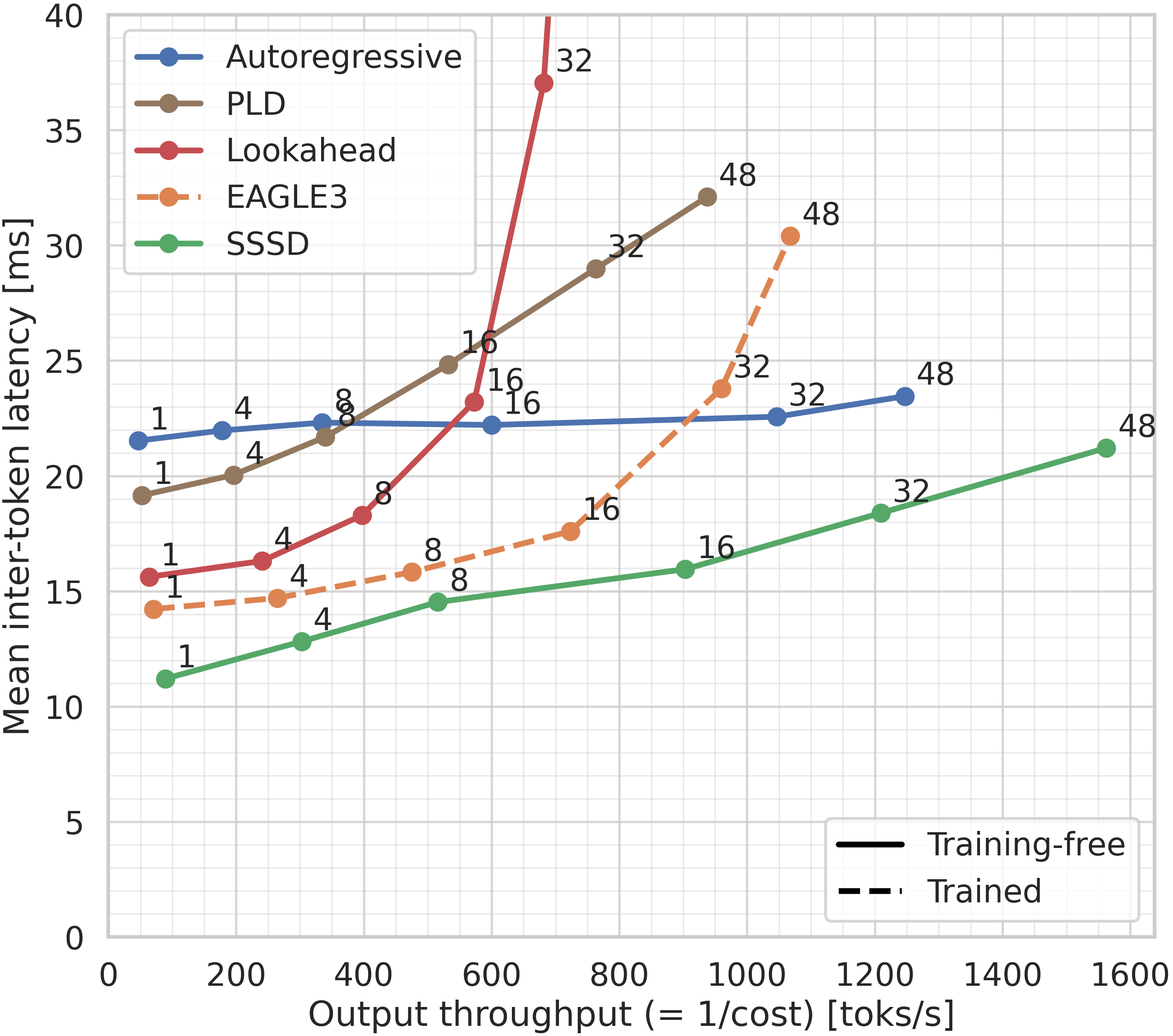}
        {\footnotesize (d) MT-Bench French}
    \end{minipage}
    \hfill
    \begin{minipage}{0.32\textwidth}
        \centering
        \includegraphics[width=\textwidth]{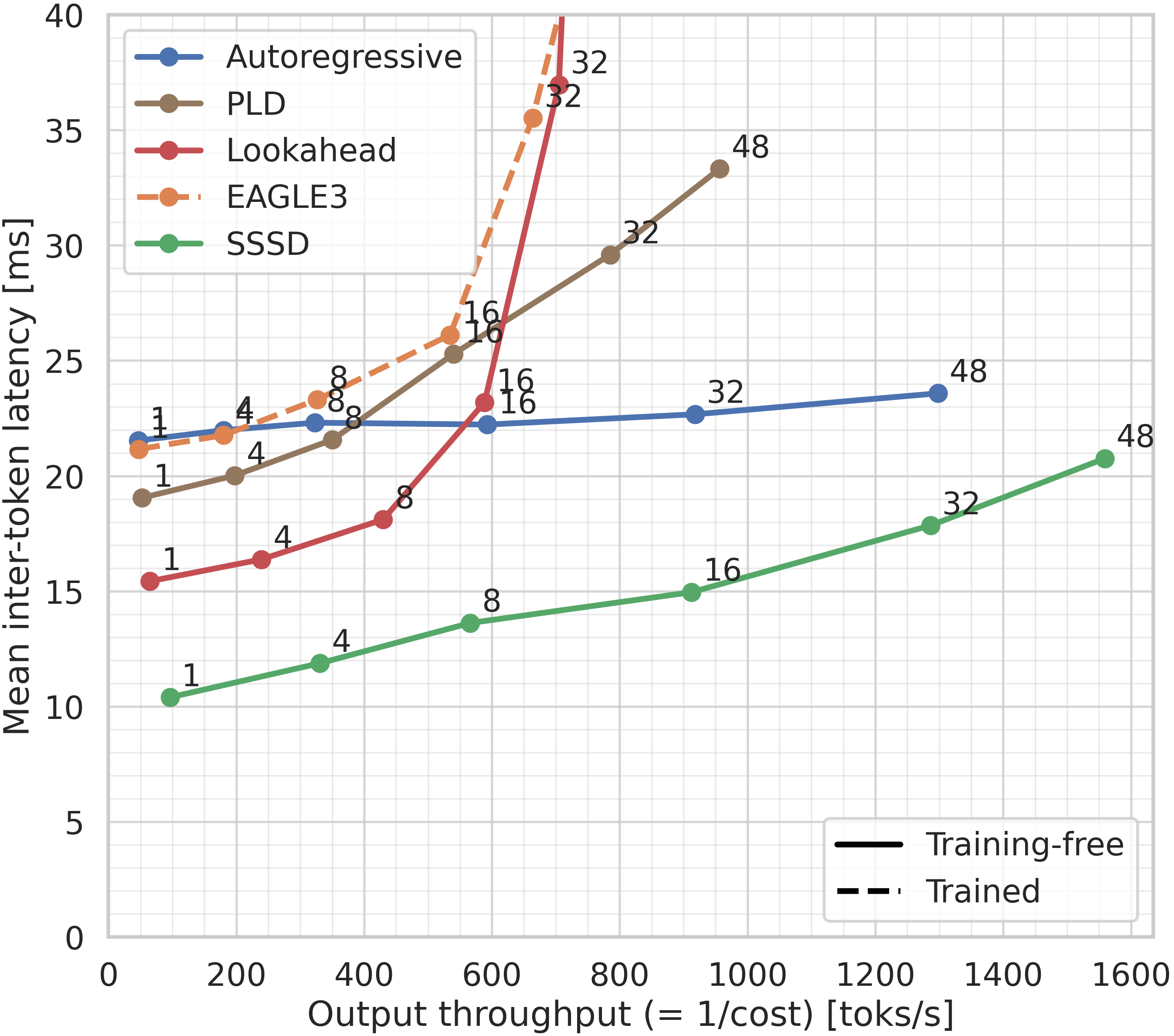}
        {\footnotesize (e) MT-Bench Russian}
    \end{minipage}
    \hfill
    \begin{minipage}{0.32\textwidth}
        \centering
        \includegraphics[width=\textwidth]{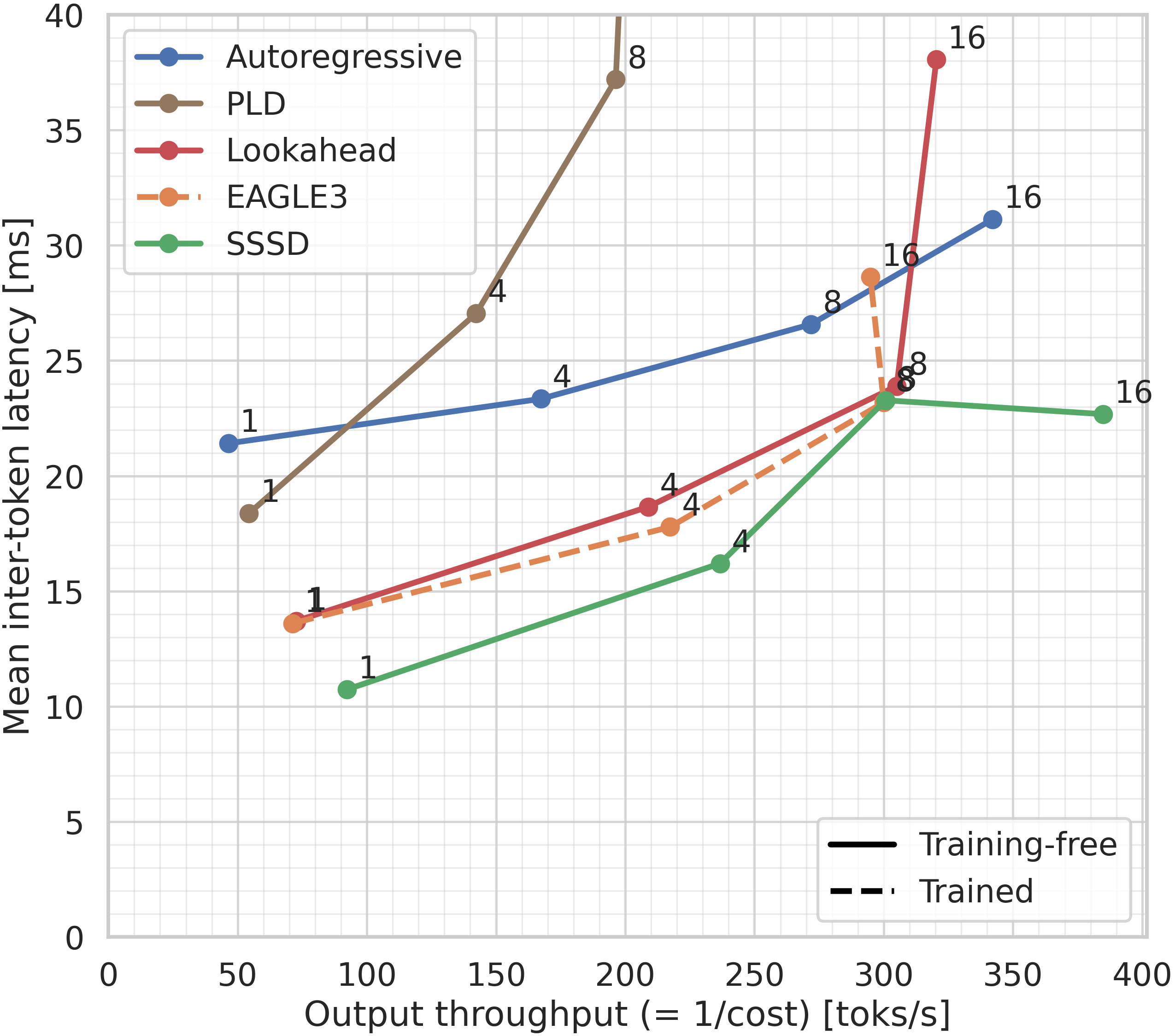}
        {\footnotesize (f) SWE-Bench}
    \end{minipage}

    \caption{Evaluation of SD methods on Llama-3.3-70B with prefill-decode disaggregation. Short context uses 2 GPUs for prefill and 2 for decode; long context uses DP=3 for prefill (6 GPUs). Prefill is overprovisioned to keep the decode node busy, though at large batch sizes throughput can be limited by prefill waiting, especially for SD methods. All GPUs: 141 GB VRAM; 989.5 TFLOPS; 4.8 TB/s. All experiments run in bfloat16 with temperature 0.}
    \label{fig:results_70b}
\end{figure*}

For the tensor-parallel (TP) setting, we ignore communication costs, which are difficult to model accurately. Although this is a coarse simplification, our empirical evaluations show that the simplified speculation length selection remains effective in practice. We leave a more detailed treatment of communication costs to future work.

\section{Large model results}
\label{app:large_model}

We additionally report results on a larger model to demonstrate that our method effectively accelerates inference even for large-scale models that require tensor parallelism. To approximate a more realistic deployment setting, we employ prefill/decode disaggregation. For short-context tasks, we use tensor parallelism with TP = 2 for both prefill and decode. For long-context tasks, we introduce three data-parallel replicas for prefill (also with TP = 2), while the decode node remains unreplicated. We further enforce a substantially larger total batch size on the prefill workers than on the decode worker, simulating a scenario in which the decode worker never stalls after receiving the initial request. To maximize throughput, we disable chunked prefill. We report performance by plotting average inter-token latency against output token throughput.

All experiments are conducted on Llama-3.3-70B. For the SSSD datastore, we use several publicly available datasets of model-generated data, resulting in a total datastore size of 13 GB. One of these datasets contains a small amount of French and Russian data; we therefore select these two languages to evaluate multilingual performance.

Our method achieves strong results across all datasets, showing consistent speedups at any batch size. On math and coding tasks, we observe speedups of 2.8× and 2.9× at batch size 1, respectively. Even in the long-context setting and for languages with limited data coverage, the speedup remains at least 2×.

Finally, we observe that on English data, for this model, EAGLE-3 outperforms SSSD. This behavior is primarily attributable to the exceptionally high acceptance rate of EAGLE-3 on this model, together with the relatively smaller size of its head compared to the base model. Despite the exceptionally performant EAGLE-3 head, SSSD demonstrates stronger adaptability to new languages and better performance in long-context scenarios.

\end{document}